\documentclass[letterpaper]{article} 
\usepackage{aaai25}  
\usepackage{times}  
\usepackage{helvet}  
\usepackage{courier}  
\usepackage[hyphens]{url}  
\usepackage{graphicx} 
\usepackage{natbib}  
\usepackage{caption} 
\usepackage{algorithm}
\usepackage{algorithmic}
\usepackage{amsmath}
\usepackage{amssymb}
\usepackage{multirow}
\usepackage{comment}
\usepackage{booktabs}
\usepackage{enumitem}

\usepackage{newfloat}
\usepackage{listings}
\DeclareCaptionStyle{ruled}{labelfont=normalfont,labelsep=colon,strut=off} 
\floatstyle{ruled}
\newfloat{listing}{tb}{lst}{}
\floatname{listing}{Listing}
\title{TIV-Diffusion: Towards Object-Centric Movement for\\ Text-driven Image to Video Generation}
\author{
    Xingrui Wang\textsuperscript{\rm 1}, Xin Li\textsuperscript{\rm 1*}, Yaosi Hu\textsuperscript{\rm 2}, Hanxin Zhu\textsuperscript{\rm 1}, Chen Hou\textsuperscript{\rm 1}, Cuiling Lan, Zhibo Chen\textsuperscript{\rm 1*}
}
\affiliations{
    \textsuperscript{\rm 1}University of Science and Technology of China, \textsuperscript{\rm 2}The Hong Kong Polytechnic University

    \{wxrui\_18264819595, hanxinzhu, houchen\}@mail.ustc.edu.cn, youncyhu@gmail.com, \\ cuilinglan@outlook.com, 
    \{xin.li, chenzhibo\}@ustc.edu.cn
}

\usepackage{bibentry}

\begin{document}
\maketitle

\renewcommand{\thefootnote}{}
\footnotetext{$^*$  Corresponding authors.}

\begin{abstract}
Text-driven Image to Video Generation (TI2V) aims to generate controllable video given the first frame and corresponding textual description. The primary challenges of this task lie in two parts: (i) how to identify the target objects and ensure the consistency between the movement trajectory and the textual description. (ii) how to improve the subjective quality of generated videos. To tackle the above challenges, we propose a new diffusion-based TI2V framework, termed TIV-Diffusion, via object-centric textual-visual alignment, intending to achieve precise control and high-quality video generation based on textual-described motion for different objects. Concretely, we enable our TIV-Diffuion model to perceive the textual-described objects and their motion trajectory by incorporating the fused textual and visual knowledge through scale-offset modulation. Moreover, to mitigate the problems of object disappearance and misaligned objects and motion, we introduce an object-centric textual-visual alignment module, which reduces the risk of misaligned objects/motion by decoupling the objects in the reference image and aligning textual features with each object individually. Based on the above innovations, our TIV-Diffusion achieves state-of-the-art high-quality video generation compared with existing TI2V methods.
\end{abstract}

%

\section{Introduction}

\label{sec:intro}
Videos that effectively convey complex visual information play a crucial role in human life, including entertainment, education, and documentation~\cite{aldausari2022VideoGANReview}. Recently, as automation advances, employing artificial intelligence algorithms for video generation has drawn numerous research attention. Early unconditional video generation~\cite{vondrick2016VGAN, saito2017TGAN, tulyakov2018mocogan} relies solely on models to learn from unlabeled video data, yielding uncontrollable outcomes. In the pursuit of controllability, approaches such as Image-to-Video Generation (I2V)~\cite{chang2021MAU,lee2021LMC-memory,gupta2022maskvit} and Text-to-Video Generation (T2V)~\cite{ho2022imagenvideo,singer2022make-a-video,wu2022tune-a-video} have come to the forefront, facilitating precise specification of appearance or movement within the generated videos. Controllable video generation enhances the way we express our intention, with diverse applications in creative content creation~\cite{blattmann2023align-your-latents}, data augmentation~\cite{wang2023modelscope}, and various fields.
\begin{figure}[!t]
   \centering
   \includegraphics[width=0.9\linewidth]{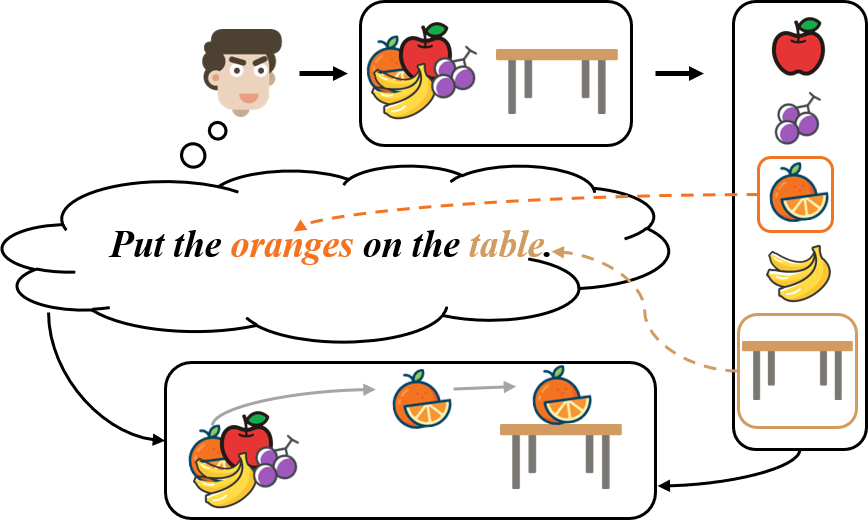}
   \caption{Humans naturally disentangle different objects in their environment. In the image above, the human wants to `Put the oranges on the table.' First, he will decouple objects with different attributes, then find the target objects according to the intention and complete the action. Our model draws inspiration from this observation.}
   \label{fig:human}
\end{figure}

Compared with images, videos introduce the temporal dimension, which means that the unconditional video generation process should ensure the authenticity of each frame as well as maintain coherence between frames~\cite{vondrick2016VGAN,sun2023moso,tulyakov2018mocogan,mei2023VIDM}. In contrast, controllable video generation is expected to consider control conditions while meeting the above-mentioned basic requirements. In particular, T2V utilizes text descriptions to specify the visual appearance and motion characteristics of the generated video~\cite{wu2021godiva,wu2022nuwa,singer2022make-a-video}. Typically, the text descriptions are encoded using CLIP~\cite{radford2021CLIP} and subsequently control the generation process via the cross-attention mechanism. While for I2V, the image or image sequence primarily restricts the visual aspects of the target video, the model should deduce subsequent object movements based on motion cues~\cite{yang2022RVD, tan2023openstl}.

Although T2V and I2V yield impressive results, it is essential to acknowledge that text inherently harbors ambiguity, leading to T2V generating numerous videos that align with a text, and I2V having limited control over motion~\cite{hu2022MAGE}. To enhance generation controllability, we direct our attention to Text-Image-to-Video Generation (TI2V), a paradigm where the image shapes the content, and the text guides movement~\cite{song2022TVP}. Nevertheless, the explorations of higher subjective quality in TI2V through diffusion models remain relatively few, and certain challenges hinder the quality of the generated videos. Firstly, given that the model processes text and image simultaneously, there emerges a necessity to align the appearance and motion of objects from different modalities. When the alignment quality is subpar, it will lead to instances where the moving objects do not correspond to the textual intended targets or their motion fails to adhere to the provided instruction~\cite{hu2022MAGE,xu2023TiV-ODE,hu2023MAGE+}. With an increased number of objects in the image, alignment difficulties are further exacerbated. Secondly, we observe that object overlap or occlusion during movement can result in deformations or disappearance of objects in subsequent video frames, as demonstrated in Sec. \ref{sec:Qualitative Results}.

\begin{figure*}[t]
    \centering
    \includegraphics[width=0.8\linewidth]{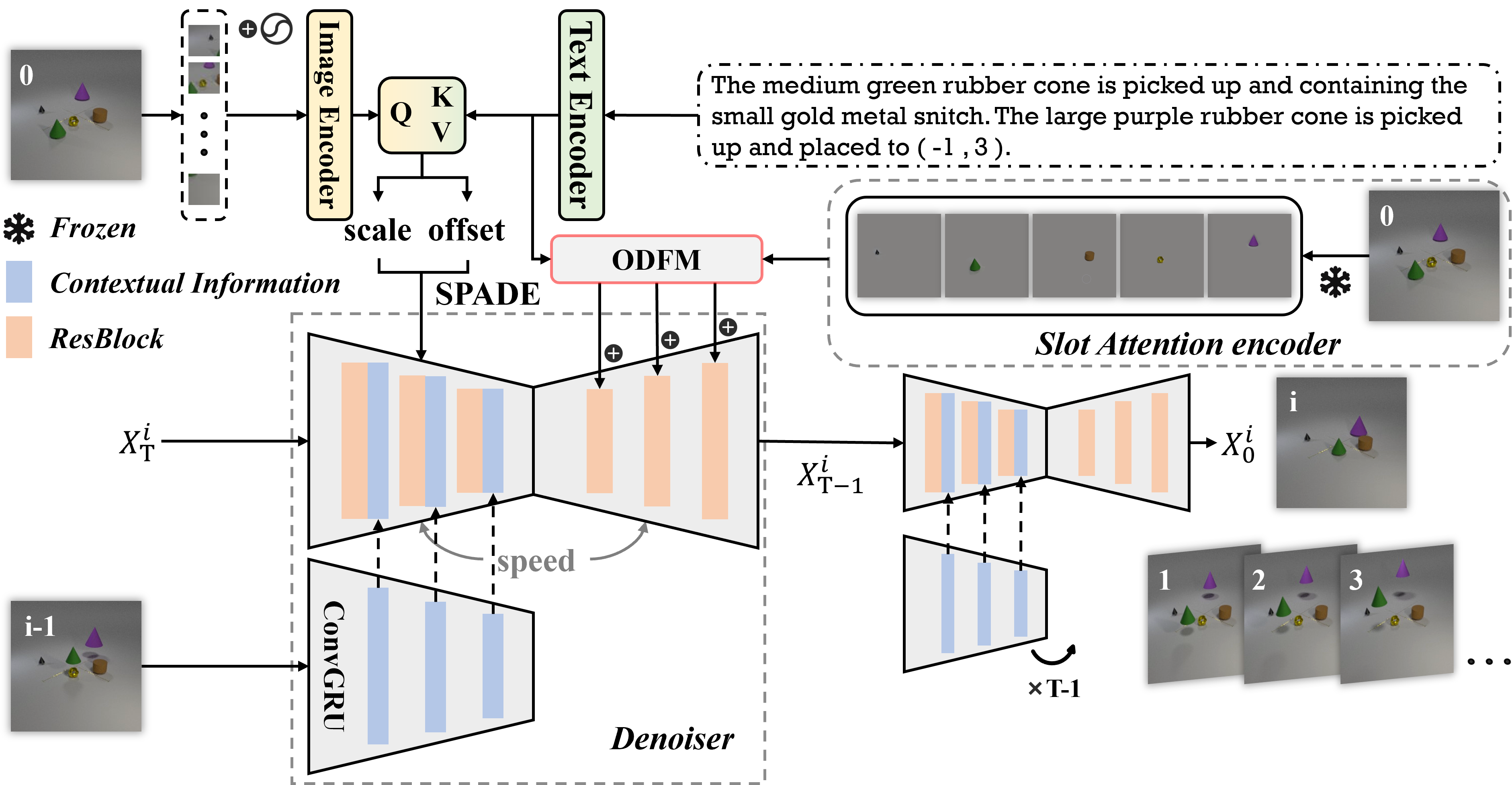}
    \caption{Illustration of the TIV-Diffusion framework. Given an image and its corresponding text caption, TIV-Diffusion can autoregressively generate subsequent video frames. To improve the alignment between objects' appearance and movement from different modalities, we introduce object-centric representations. We employ a slot attention encoder to extract slots and further fuse them with text information through the Object Distanglement Fusion Module (ODFM), which will be introduced thoroughly in Sec. \ref{sec:Utilize Object Disentanglement for Improved Alignment}. Additionally, the video generation process adaptively incorporates object attribute information from slots to address object deformation and disappearance.}
    \label{fig:framework}
\end{figure*}

As illustrated in Fig.~\ref{fig:human}, motivated by the natural ability of individuals to decouple various objects within their visual field and correlate them with the corresponding text description, we propose a new framework for \textbf{TI}2\textbf{V}, dubbed \textbf{TIV-Diffusion}, leveraging object disentanglement to improve textual-visual alignment and subjective quality.
Specifically, as shown in Fig.~\ref{fig:framework}, TIV-Diffusion encodes the input image and text caption respectively, followed by a fusion of their encoded embeddings, and then modulates them into the autoregressive generation process in a SPADE~\cite{park2019SPADE} manner, which integrates the appearance and motion information.
Furthermore, TIV-Diffusion extracts object-centric representations (\textit{i.e.}, slots) from the input image utilizing a Slot Attention~\cite{locatello2020SlotAttention} encoder. This encoder aims to discover the latent compositional structure from unstructured observations~\cite{jiang2023Object_Centric_Slot_Diffusion}, and the resulting slots capture object attributes. To facilitate TIV-Diffusion in comprehending which objects to move and their final destinations, as depicted in Fig.~\ref{fig:gumbel}, we align slots with the text caption individually.

Disentangling objects enables precise alignment between objects and their motion description, fostering a heightened semantic consistency between the generated video and its corresponding text. Moreover, to mitigate object disappearance or deformation, we incorporate slots into the generation process. Hence, with the object attributes in slots, TIV-Diffusion refines object composition by constant awareness of object features. However, considering that the object stored in each slot is unknown, we use Gumbel-Softmax~\cite{jang2016Gumbel-Softmax} for adaptive slot selection, thereby improving the temporal consistency of objects.

The contributions of this paper are concluded as follows:
\begin{itemize}
  \item We introduce a diffusion-based TI2V model that harnesses the potential of object disentanglement, which can generate video frames with high perceptual quality, as well as achieve better controllability.
  \item Our model extracts object features (\textit{i.e.}, slots) through the Slot Attention encoder to enhance the alignment between text and objects, and incorporates slots during the generation process adaptively to mitigate the likelihood of object deformation and disappearance.
  \item We perform experiments on two categories of existing datasets, and the model performs well under various control conditions. Extensive results have demonstrated that our proposed method can achieve state-of-the-art performance.
\end{itemize}

\section{Related Work}
\label{sec:Related Work}
\paragraph{Text-Image-to-Video Generation.} Text-Image-to-Video Generation achieves enhanced controllability through combining image and text. TVP~\cite{song2022TVP} explores the causality in the text description and then generates step-wise inference embeddings to guide the generation of each frame using GAN~\cite{goodfellow2014GAN}. Concurrent work MAGE~\cite{hu2022MAGE} employs VQVAE for frame quantization and leverages a transformer-based approach guided by a motion anchor. Furthermore, the MAGE+~\cite{hu2023MAGE+} variant enhances performance by incorporating a robust compression autoencoder. MMVG~\cite{fu2023MMVG} is similar to the former, except that it combines masks to achieve multi-tasks. TiV-ODE~\cite{xu2023TiV-ODE} considers temporal continuity and implements the Neural ODE approach, but the generated videos have residual shadows. LFDM~\cite{ni2023LFDM} utilizes diffusion models to learn the distribution of optical flow sequences, which are used to synthesize single-object videos. Similar to the motivation of TVP, diffusion-based Seer~\cite{gu2023seer} uses fine-grained textual sub-instructions for each frame generation. DynamiCrafter~\cite{xing2023dynamicrafter}, on the other hand, incurs expensive computational costs.

One primary challenge in the TI2V task lies in aligning the text caption and the image. When misaligned, object movement trajectories will be semantically inconsistent with the text description. Unlike prior works, we enhance this alignment by introducing object-centric features and achieve fine-grained control in a resource-friendly manner.
\paragraph{Other Controllable Video Generation.} Text-to-Video Generation refers to the generation of videos guided by textual input. GODIVA~\cite{wu2021godiva}, CogVideo~\cite{hong2022cogvideo}, and NÜWA~\cite{wu2022nuwa} leverage codebooks from VQVAE or VQGAN for video content quantization. They employ sparse attention mechanisms to capture textual information and temporal relationships within the video efficiently. Meanwhile, diffusion models excel in image generation and have recently been extended to the realm of video generation~\cite{singer2022make-a-video,zhou2022magicvideo,ho2022imagenvideo,wu2022tune-a-video,an2023latent-shift,wang2024cono}. To adapt U-Net for video data, current works often incorporate temporal convolutions or temporal attention mechanisms while using cross-attention to introduce text guidance into the generation process~\cite{singer2022make-a-video,ho2022imagenvideo,zhou2022magicvideo}. Besides, some works choose to generate videos in the latent space, utilizing pre-trained Stable Diffusion~\cite{rombach2022StableDiffusion} model weights for initialization, and exclusively fine-tuning temporal layers for efficient training~\cite{an2023latent-shift,blattmann2023align-your-latents,xing2023simda}.

Video-to-Video Generation focuses on predicting future video frames using motion cues from past frames. Recurrent-based models, including ConvLSTM~\cite{shi2015ConvLSTM} and ConvGRU~\cite{ballas2015ConvGRU} structures, are extensively studied in this context. LMC-memory~\cite{lee2021LMC-memory} employs memory to store long-term motion context and leverages this information to assist ConvLSTM in predicting future frames, while MAU~\cite{chang2021MAU} introduces a motion-aware unit to expand the temporal receptive field. Compared to recurrent-based models, recurrent-free models~\cite{tan2023TAU,tan2023openstl}, which process multiple frames concurrently, offer a trade-off between efficiency and performance. Furthermore, the Transformer~\cite{vaswani2017transformer} architecture is also a promising option \cite{weissenborn2019ScalingAutoregressive,he2022latent-video-diffusion,gupta2022maskvit,sun2023moso}. Diffusion models, exemplified by MCVD~\cite{voleti2022mcvd} and RaMViD~\cite{hoppe2022RaMViD}, excel in multi-tasks such as frame prediction and infilling. Unlike conventional methods that directly forecast video frames, RVD~\cite{yang2022RVD} chooses to predict frame residuals.

\section{Method}
\label{sec:Method}
\subsection{Preliminary}
\label{sec:Preliminary}
\textbf{Diffusion models}~\cite{ho2020DDPM,song2020DDIM,li2023diffusion,ren2025moe} primarily acquire knowledge of the unknown data distribution through the dual processes of \emph{diffusion} and \emph{denoising}.
\emph{Diffusion} gradually disrupts data distribution with Gaussian noise, avoiding extra parameter training.
This process can be represented as a Markov chain denoted by $q\left( X_t \mid X_{t-1} \right) =\mathcal{N} \left( X_t \mid \sqrt{1-\beta _t}X_{t-1},\beta _t\mathrm{I} \right)$, where $\beta_t\in(0,1)$ and $X_t$ represent the $t$th step of noise addition, with a total of \emph{T} steps.
\emph{Denoising} is used for data structure reconstruction, and in DDPM~\cite{ho2020DDPM}, $\epsilon$-prediction is introduced.
This method involves learning the denoising function $f_{\theta}$ through mean square error loss minimization, represented as $L\left( \theta \right) =\mathbb{E} _{X_0,t,\epsilon}\left\| \epsilon -f_{\theta}\left( X_t,t \right) \right\| ^2$. Following model training, desired data samples can be acquired by sampling Gaussian noise and continuously applying denoising processes.
\subsection{Overall Framework}
Given an image $X^0$ along with a text caption $S=\left\{ s_1,\cdots ,s_L \right\}$ of length $L$, the model aims to generate a new sequence of images $X^{1:N}=\left\{ X^1,\cdots ,X^N \right\}$. These generated images should maintain visual consistency with $X^0$ while adhering to the motion described in $S$, represented by the conditional distribution $p\left( X^{1:N}\left| X^0,S \right. \right)$.

Fig.~\ref{fig:framework} illustrates the overall architecture of the proposed TIV-Diffusion. Our method is founded on the diffusion model, which allows for controllable video generation through extensions discussed in Sec.~\ref{sec:Autoregressive TI2V Generation}. Previous works lack disentanglement of objects within the image, leading to suboptimal alignment between the text caption and the image. In Sec. \ref{sec:Utilize Object Disentanglement for Improved Alignment}, we incorporate object-centric representations to enhance this alignment.
\subsection{Autoregressive TI2V Generation}
\label{sec:Autoregressive TI2V Generation}

To reduce computational demand, we employ an autoregressive generation approach here~\cite{voleti2022mcvd,yang2022RVD}.
\paragraph{Text-Image Fusion.}

The text encoder, comprised of trainable Transformer~\cite{vaswani2017transformer} encoder layers, establishes correlations within the input data while encoding the text $S$ as $e_S\in \mathbb{R} ^{L\times d}$, where $d$ denotes the dimension of the text embeddings. The image, denoted as $X^0\in \mathbb{R} ^{H\times W\times C}$, is partitioned into non-overlapping patches of uniform size~\cite{dosovitskiy2020ViT}, where $H$ and $W$ represent the height and width, and $C$ is the number of channels. The image token is also encoded using Transformer encoder layers and position embeddings are learned during training. Since each patch has a size of $f_p\times f_p$, the encoding of $X^0$ is represented as $e_{X^0} \in \mathbb{R}^{\frac{H}{f_p} \times \frac{W}{f_p} \times d}$. Following this, information from both modalities is fused through cross-attention, with $e_{X^0}$ as the Query and $e_S$ as the Key and Value:
\begin{equation}
    c=\mathrm{CrossAttention}\left( Q\left( e_{X^0} \right) ,K\left( e_S \right) ,V\left( e_S \right) \right).
\end{equation}

Here, $Q$, $K$, and $V$ are learnable linear projections. The fused information $c$ aligns the textual description of objects with their respective counterparts in the image, specifying their motion.
$c$ will be incorporated into the downsampling section of Unet~\cite{ronneberger2015Unet} in the SPADE~\cite{park2019SPADE} format, where it is utilized for calculating multi-scale scaling and offset values to modulate the feature maps $\left\{ \mathbf{F}_{\mathrm{Unet}}^{m} \right\} _{m=1}^{M}$ within the residual blocks of Diffusion. $M$ represents the number of down or up-sampling layers.
\begin{equation}
\label{eq:SPADE}
\begin{split}
    \mathbf{\hat{F}}_{\mathrm{Unet}}^{m}=&\left( 1+\gamma ^m \right) \odot \mathbf{F}_{\mathrm{Unet}}^{m}+\beta ^m,\\
&\gamma ^m,\beta ^m=\mathcal{M} _{\theta}^{m}\left( c \right) .
\end{split}
\end{equation}
In Eq. \ref{eq:SPADE}, the scale $\gamma ^m$ and offset $\beta ^m$ are the modulation parameters, $\mathcal{M} _{\theta}^{m}$ encompasses multiple convolutional layers and activation functions, and $m$ denotes the current layer. In the \textbf{Supplementary Material~\ref{sec:Further analysis on SPADE}}, we further explore the impact of $\gamma ^m$ and $\beta ^m$ on object movement.
\paragraph{Text-Image-to-Video Generation.} 
The video generator, an extension of DDPM~\cite{ho2020DDPM}, integrates a timing relationship capture module. Specifically, it conditions the generation of video frame $X^n$ on $X^{<n}$ (all frames before $n$), and information extraction from $X^{<n}$ is achieved iteratively, using ConvGRU~\cite{ballas2015ConvGRU}.
Initially, $X^0$ is processed through ConvGRU to extract contextual information, and then concatenated with $\left\{ \mathbf{\hat{F}}_{\mathrm{Unet}}^{m} \right\} _{m=1}^{M}$ at multiple scales to produce the video frame $X^1$.
Similarly, $X^1$ is used as a condition to generate $X^2$ by passing it through ConvGRU for status updates.
Continuing this process, each $X^n$ generation involves sending $X^{n-1}$ to the timing module to gather dynamic information from frame $0$ to $n-1$.
With the concurrent guidance from $c$ and $X^{<n}$, we employ frame reconstruction loss~\cite{zhou2022magicvideo} directly to train the diffusion model:
\begin{equation}
\label{eq:diffusion_loss_speed}
\begin{split}
    \mathcal{L} \left( \theta \right) =\mathbb{E} _t\sum_{n=1}^N&{\left\| X_{0}^{n}-f_{\theta}\left( X_{t}^{n}|t,c,X^{<n},v \right) \right\| _{2}^{2}},\\
    &v=\phi \left( \eta \right) .
\end{split}
\end{equation}
To enhance video diversity, we control objects' speed using the parameter $\eta$, encoding it with a linear layer $\phi$, and introduce it via cross-attention.

\subsection{Utilize Object Disentanglement for Improved Alignment}
\label{sec:Utilize Object Disentanglement for Improved Alignment}
\begin{figure}[!h]
   \centering
   \includegraphics[width=1.0\linewidth]{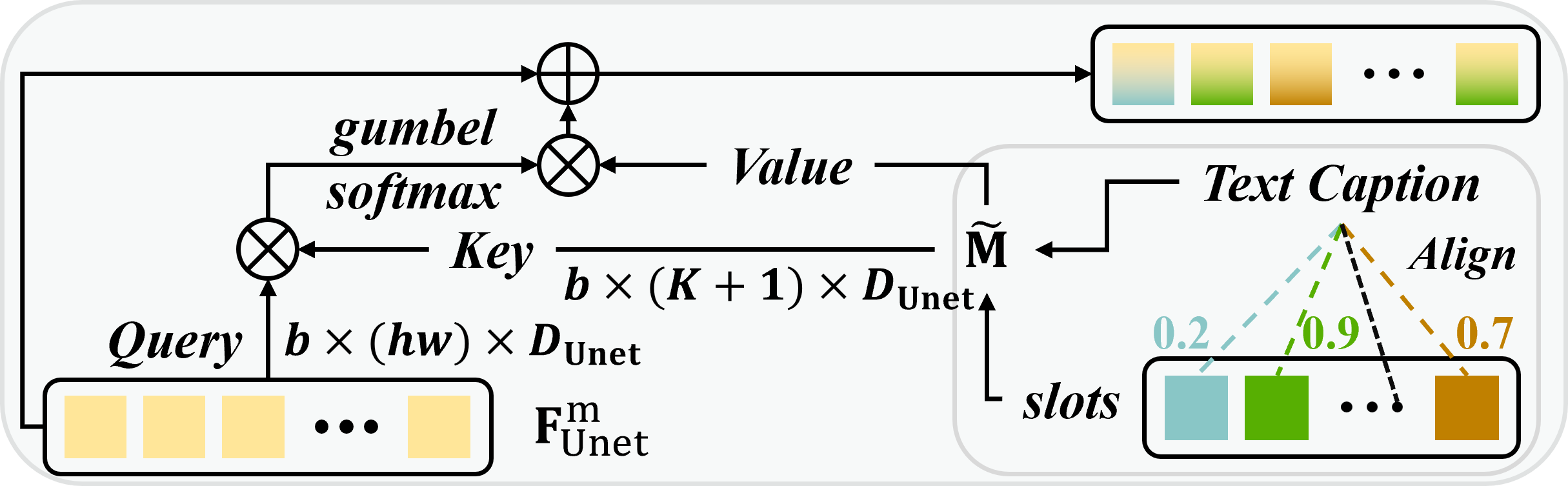}
   \caption{Object Distanglement Fusion Module (ODFM). We initially identify the relevant target moving objects within the text caption based on the provided slots and then align slots with the text caption individually. Subsequently, we utilize the text-enhanced slots to facilitate the video generation process.}
   \label{fig:gumbel}
\end{figure}

Assuming there are $K$ objects in the initial image $X^0$, along with the background, a total of $K+1$ corresponding learnable variables are required. We utilize the Slot Attention~\cite{locatello2020SlotAttention} encoder to extract object features from $X^0$, as shown in Fig. \ref{fig:framework}. Specifically, the $K+1$ slots, each possessing dimension $D_{\mathrm{slots}}$, are initialized using a Gaussian distribution featuring the learnable $\mu$ and $\sigma$, denoted as $\mathrm{slots}\in \mathbb{R} ^{\left( K+1 \right) \times D_{\mathrm{slots}}}$. Given an image $X^0$ and processed by CNNs~\cite{lecun1989CNNs}, we get $\mathrm{inputs}\in \mathbb{R} ^{N_{\mathrm{inputs}}\times D_{\mathrm{inputs}}}$. Subsequently, we employ Scaled Dot-Product Attention~\cite{vaswani2017transformer} $\mathrm{Softmax} \left( \frac{1}{\sqrt{D}}k\left( \mathrm{inputs} \right) \cdot q\left( \mathrm{slots} \right)^T,\mathrm{axis}=``\mathrm{slots}"\right)$ to foster competition among slots and iteratively update the content within them, thereby yielding object-centric representations. Before that, we should apply learnable linear transformations to map $D_{\mathrm{inputs}}$ and $D_{\mathrm{slots}}$ to a shared dimension $D$. The Slot Attention encoder is trained via image reconstruction: 
\begin{equation}
    \hat{X}^0=Dec\left( Enc\left( X^0 \right) \right).
\end{equation}

$Enc$ (\textit{i.e.}, Slot Attention encoder) in this context encompasses both the CNNs responsible for feature extraction from $X^0$ and the slots update module, while $Dec$ uses slots to reconstruct $X^0$~\cite{locatello2020SlotAttention}. $Enc$ and $Dec$ are pre-trained before training the TI2V diffusion model. $Dec$ is a CNN-structured auxiliary module designed to aid in the training of $Enc$, and it is omitted in Fig. \ref{fig:framework}.
Following this, object-centric representations are obtained using Eq. \ref{eq:slots}:
\begin{equation}
\label{eq:slots}
    \mathrm{slots}=Enc\left( X^0 \right).
\end{equation}

Fig.~\ref{fig:gumbel} shows how object slots can be used to improve the effects of the generated video. The input text caption specifies the objects to be controlled. As a result, the currently obtained slots make it easier to extract relevant information from the text, improving the identification of the target objects. Additionally, slots also store object attribute information.
Combining the above two aspects, we get Eq. \ref{eq:SlotsAndText}:
\begin{equation}
\label{eq:SlotsAndText}
    M=\mathrm{slots}+\mathrm{CrossAttention}\left( Q\left( \mathrm{slots} \right) ,K\left( e_S \right) ,V\left( e_S \right) \right). 
\end{equation}
We align slots with the text caption individually, and the slots augmented with textual information are denoted as $M\in \mathbb{R} ^{\left( K+1 \right) \times D}$.
The experiments in Sec.~\ref{sec:Experiments} demonstrate that $M$ effectively enhances text-image alignment and mitigates object deformation.
Relying on similarity within the embedding space, we enable the diffusion model to autonomously select suitable slots, thereby enhancing the generation of video frames.
In order to match the inner dimension $D_{\mathrm{Unet}}$ of Unet, we first linearly project $M$ to $\tilde{M}=\mathrm{Linear}\left( M \right) \in \mathbb{R} ^{\left( K+1 \right) \times D_{\mathrm{Unet}}}$.
Gumbel-Softmax~\cite{jang2016Gumbel-Softmax} is used for interaction between $\mathbf{F}_{\mathrm{Unet}}^{m}$ and $\tilde{M}$ to introduce randomness and allow the model to explore during the training process.
Due to the unpredictable assignment of objects to slots, this exploration process increases model diversity to handle uncertainty effectively.
\begin{equation}
\label{eq:GumbelSoftmax}
    A_{i,j}^{m}=\frac{\exp \left( W_q\mathbf{F}_{\mathrm{Unet},i}^{m}\cdot W_k\tilde{M}_{j}^{m}+\varepsilon _j \right)}{\sum\nolimits_{k'=1}^{K+1}{\exp \left( W_q\mathbf{F}_{\mathrm{Unet},i}^{m}\cdot W_k\tilde{M}_{k'}^{m}+\varepsilon _{k'} \right)}}
\end{equation}
In practice, we introduce $\tilde{M}$ using Eq.~\ref{eq:GumbelSoftmax} after each resolution's ResBlock in the upsampling section of Unet. $W_q$ and $W_k$ are the weights of learned linear projections, and $\left\{ \varepsilon _j \right\}$ are i.i.d. random samples from the $\mathrm{Gumbel(}0,1)$ distribution. We empower the model to adaptively choose the required object features by conducting the $\mathrm{argmax}$ operation across all acquired slots, and further employ a straight-through trick to address the issue of the $\mathrm{argmax}$ operation being non-differentiable~\cite{xu2022groupvit,van2017NeuralDiscrete}:
\begin{equation}
\label{eq:straight-through}
    \hat{A}^m=\mathrm{one-hot}\left( A_{\mathrm{argmax}}^{m} \right) +A^m-\mathrm{sg}\left( A^m \right).
\end{equation}
Furthermore, we incorporate them into the feature maps of the original Unet using a residual approach, as illustrated in Fig. \ref{fig:framework}:
\begin{equation}
\label{eq:ResAtten}
    \mathbf{F}_{\mathrm{Unet}}^{m+1}=\mathbf{F}_{\mathrm{Unet}}^{m}+\frac{\sum\nolimits_{j=1}^{K+1}{\hat{A}_{i,j}^{m}}W_v\tilde{M}_{j}^{m}}{\sum\nolimits_{j=1}^{K+1}{\hat{A}_{i,j}^{m}}}.
\end{equation}
Finally, Eq. \ref{eq:diffusion_loss_speed} is modified to:
\begin{equation}
\label{eq:GumbelSoftmaxDiffusion}
    \mathcal{L} \left( \theta \right) =\mathbb{E} _t\sum_{n=1}^N{\left\| X_{0}^{n}-f_{\theta}\left( X_{t}^{n}|t,c,X^{<n},v,\tilde{M} \right) \right\| _{2}^{2}}.
\end{equation}
During training, the parameters of $Enc$ remain fixed.

\section{Experiments}
\label{sec:Experiments}
\begin{figure}[!h]
   \centering
   \includegraphics[width=1.0\linewidth]{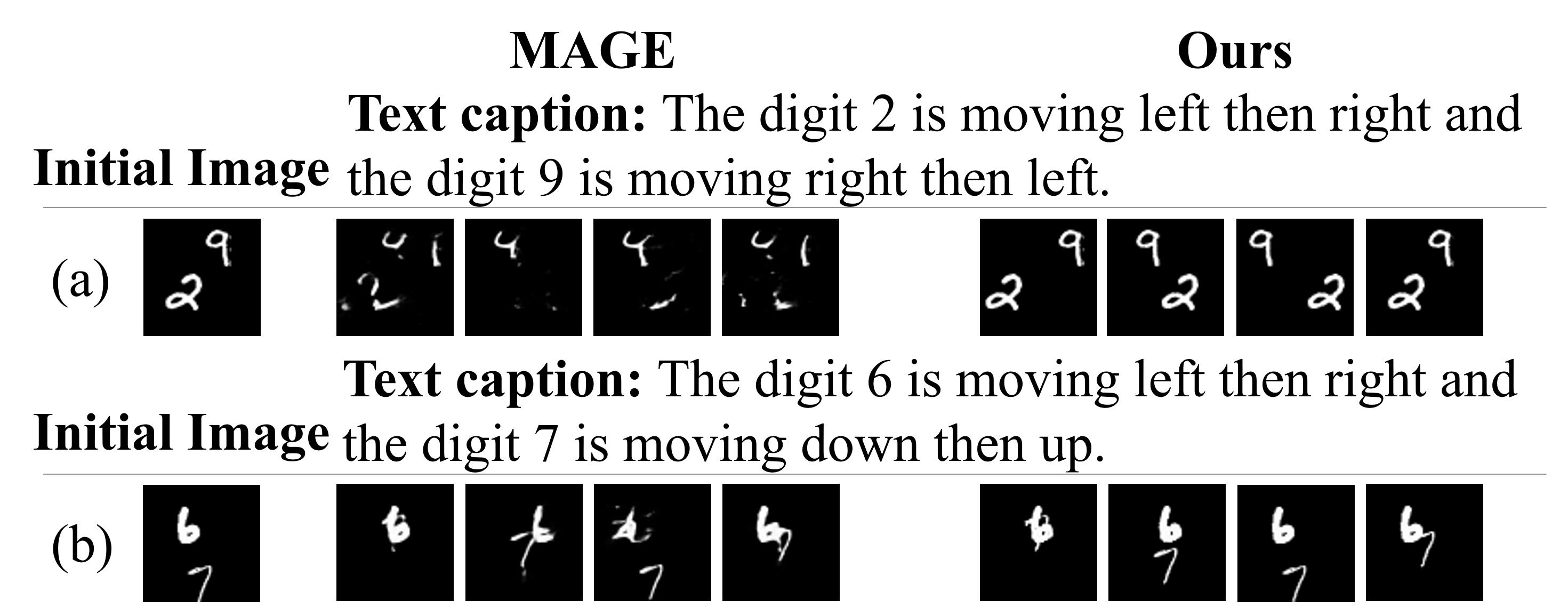}
   \caption{Comparison Results on Double Moving MNIST. Under the constraints of object-centric representations, the shapes of the digits in the videos generated by our model remain consistent even when they overlap. To facilitate visualization, we have extracted specific frames.}
   \label{fig:VS_MNIST}
\end{figure}
\begin{figure*}[t]
    \centering
    \includegraphics[width=0.8\linewidth]
    {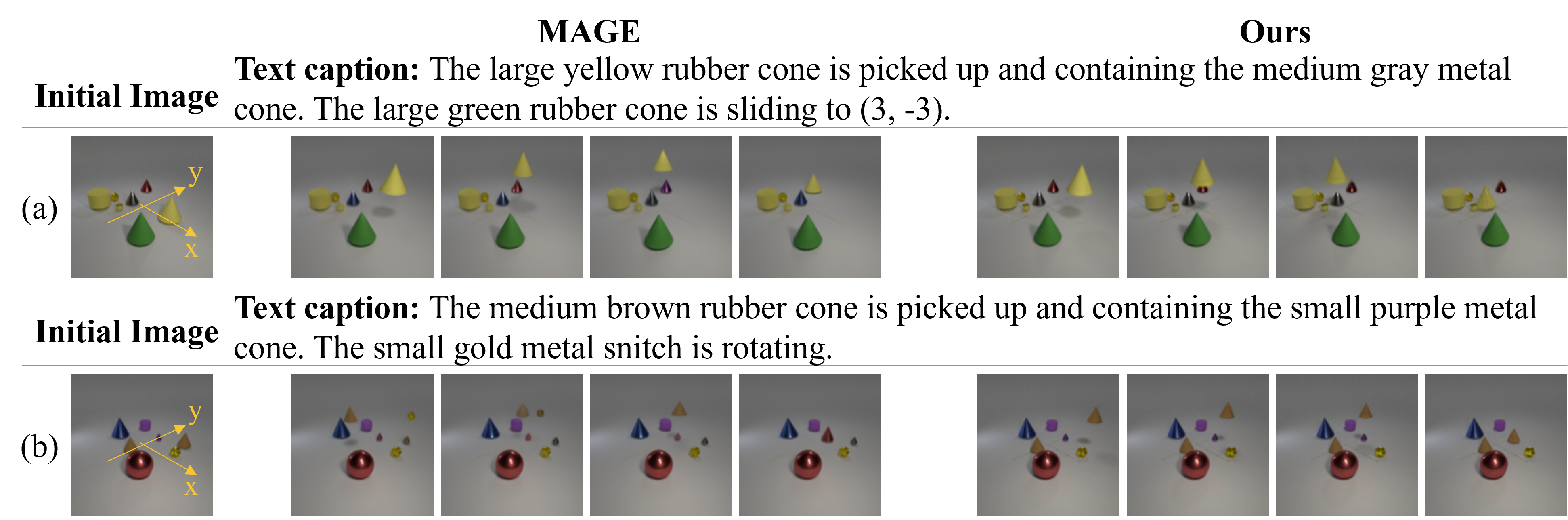}
    \caption{Comparison Results on CATER-GEN-v2. We propose object disentanglement for enhanced alignment between text and image in video generation. This yields improved semantic consistency, enabling precise identification and motion in accordance with textual descriptions.}
    \label{fig:VS_CATER2_Align}
\end{figure*}
\begin{figure*}[t]
    \centering
    \includegraphics[width=0.8\linewidth]
    {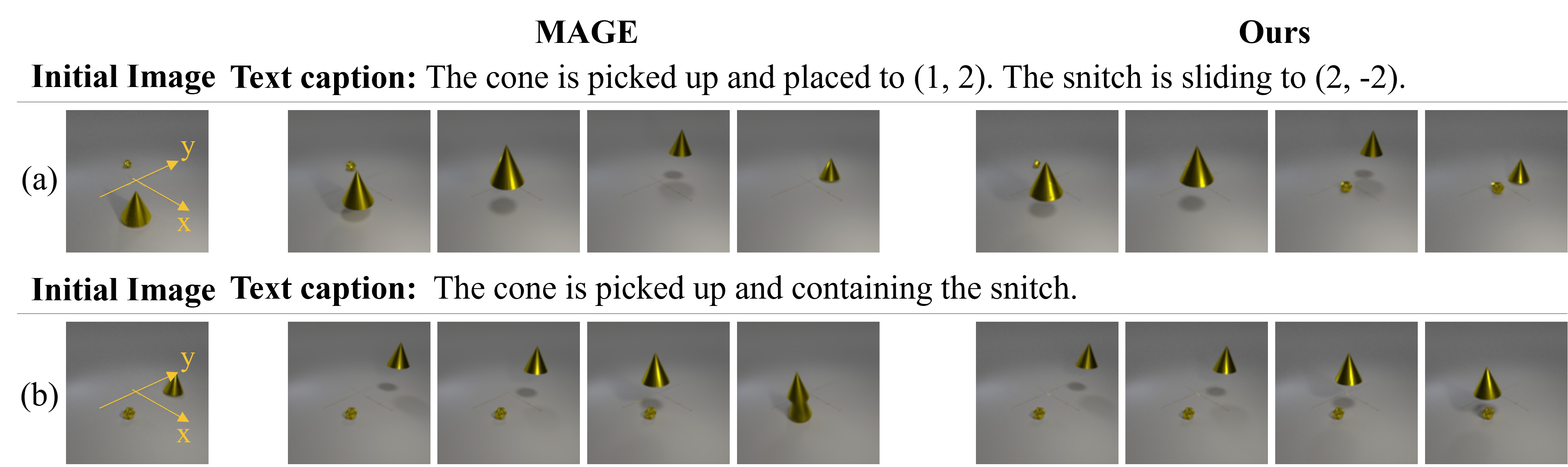}
    \caption{Comparison Results on CATER-GEN-v1. We can observe the object's disappearance or deformation as the video continues. Instead, our model alleviates it by utilizing the object attribute information in slots to improve the generation process.}
    \label{fig:VS_CATER}
\end{figure*}
\begin{figure*}[t]
    \centering
    \includegraphics[width=0.8\linewidth]
    {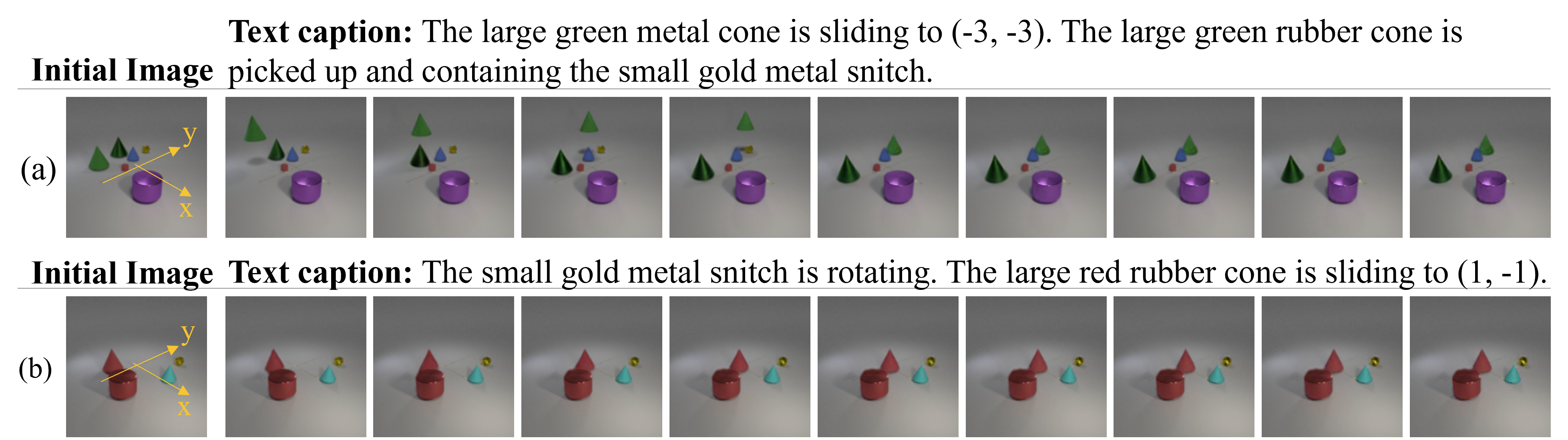}
    \caption{Samples generated from CATER-GEN-v2. The initial image on the far left is provided by the user, and above the images is a text caption describing the movement. The generated videos are coherent and of high perceptual quality, maintaining semantic consistency with textual descriptions.}
    \label{fig:caterv2}
\end{figure*}
\subsection{Datasets and Evaluation Metrics}
\label{sec:datasets}
MAGE~\cite{hu2022MAGE} introduces five datasets for evaluating this task, comprising three MNIST datasets and two CATER datasets. Additionally, to assess the performance of our model in more realistic scenarios, we select the Weizmann dataset~\cite{gorelick2007actions}, KTH dataset~\cite{schuldt2004kth}, and portions of the Bridge dataset~\cite{ebert2021bridge} for testing. Due to space limitations, the relevant results are presented in the \textbf{Supplementary Material~\ref{sec:weizmann results}}, \textbf{\ref{sec:kth results}}, and \textbf{\ref{sec:bridge results}}.
\paragraph{MNIST datasets.} Single Moving MNIST contains a single digit, whereas Double Moving MNIST~\cite{mittal2017Sync-draw} features pairs of digits moving in various directions: top to bottom, bottom to top, left to right, and right to left.
Modified Double Moving MNIST differs in movement, with digits stopping or bouncing once at boundaries and a random static digit inserted in the background.
It's worth noting that we slightly modify these three datasets because an examination of the source code reveals that the digits exhibit uneven speeds when reaching the boundaries and rebounding, which affects the evaluation under varying speed conditions.
Consequently, these datasets are adjusted to ensure uniform motion.
The video resolution is $64 \times 64$ pixels.
\paragraph{CATER datasets.} MAGE~\cite{hu2022MAGE} provides two datasets, namely CATER-GEN-v1 and CATER-GEN-v2, which remain unaltered in this paper.
CATER-GEN-v1 comprises two objects (a cone and a snitch), while CATER-GEN-v2 encompasses 3 to 8 objects, each defined by four randomly selected attributes like shape and material.
The atomic actions of objects in these two datasets are the same, namely, ``rotate'', ``contain'', ``pick-place'', and ``slide''.
The coordinate system in which the objects move is shown in the \textbf{Supplementary Material~\ref{app:coordinate system}}. The video resolution is $256 \times 256$, but it will be resized to $128 \times 128$ during the experiments.

In accordance with prior work~\cite{xu2023TiV-ODE,hu2022MAGE}, we select evaluation metrics including Peak Signal to Noise Ratio (PSNR)~\cite{huynh2008PSNR}, Structural Similarity Index Measure (SSIM)~\cite{wang2004SSIM}, Learned Perceptual Image Patch Similarity (LPIPS)~\cite{dosovitskiy2016LPIPS}, Fréchet Inception Distance (FID)~\cite{heusel2017FID} and Fréchet-Video-Distance (FVD)~\cite{unterthiner2018FVD}.
In the computation of LPIPS, the VGG~\cite{simonyan2014VGG} network was employed.
\subsection{Qualitative Results}
\label{sec:Qualitative Results}
This section presents the qualitative results on the datasets mentioned in Sec. \ref{sec:datasets}.
\paragraph{Comparison Results.} In Figs. \ref{fig:VS_MNIST}, \ref{fig:VS_CATER2_Align} and \ref{fig:VS_CATER}, we show video generation results comparing our model with MAGE~\cite{hu2022MAGE}. In cases of text-image misalignment, two scenarios emerge: firstly, the model fails to align object motion with the textual description, and secondly, it mistakenly identifies the objects to be moved. Fig.~\ref{fig:VS_CATER2_Align}(a) illustrates the first scenario, wherein MAGE successfully identifies the large yellow rubber cone but struggles to execute the action containing the medium gray metal cone. Fig.~\ref{fig:VS_CATER2_Align}(b) corresponds to the second scenario, where MAGE misidentifies the medium brown rubber cone requiring movement. Our model employs object disentanglement to separate co-located objects, storing object attribute information in discrete slots. This strategy facilitates the alignment of textual descriptions, resulting in improved generation results. 

As the number of frames increases, objects in MAGE-generated video frames deform. Instead, our approach utilizes object-centric representations to enhance the generation process, thus ensuring constant awareness of object attributes. This is evident in the comparison results in Fig.~\ref{fig:VS_MNIST} and Fig.~\ref{fig:VS_CATER}(b). When objects overlap, as observed in Fig.~\ref{fig:VS_CATER}(a) of the MAGE generation results, one of the objects may disappear. However, with the assistance of object features, our method improves the temporal consistency of objects.
\paragraph{Generated Results.} The CATER-GEN-v2 dataset poses significant challenges due to varying object sizes and multiple objects with identical shapes, demanding a high level of discrimination capability from the model. In Fig. \ref{fig:caterv2}(a), there are cones exhibiting distinct attributes, including two cones of green color. Distinguishing between these two cones relies on the model's analysis of their unique material properties. It can be observed that TIV-Diffusion successfully recognizes the difference between metal and rubber, executing the correct motion instructions. The generated results on the MNIST datasets can be found in \textbf{Supplementary Material~\ref{sec:Generated Results on MNIST}}.

We provide additional details on the effects relevant to controllability evaluation and more generated results (\textit{e.g.}, Seer) in the \textbf{Supplementary Material~\ref{sec:additional qualitative results}}.
\begin{table}[!h]
\centering
\small
\setlength{\tabcolsep}{1mm}
\begin{tabular}{@{}cl|ccccc@{}}
\toprule
Datasets                                      & Method          & SSIM$\uparrow$ & PSNR$\uparrow$ & LPIPS$\downarrow$ & FID$\downarrow$  & FVD$\downarrow$\\ \midrule
\multirow{3}{*}{Single}                       & TVP             & 0.46  & 17.24 & 0.30  & - & - \\
                                              & MAGE$^{\dagger}$& 0.98  & 32.98 & 0.02  & 5.94 & 1.20\\
                                              & \textbf{Ours}   & \textbf{0.99} & \textbf{36.68} & \textbf{0.01} & \textbf{0.49} & \textbf{0.11}\\ \midrule
\multirow{2}{*}{Double}                       & MAGE$^{\dagger}$     & 0.92  & 26.13 & 0.06  & 7.83 & 2.55\\
                                              & \textbf{Ours}   & \textbf{0.96} & \textbf{32.99} & \textbf{0.03} & \textbf{1.04} & \textbf{0.47}\\ \midrule
\multirow{2}{*}{Modified}                     & MAGE$^{\dagger}$   & \textbf{0.92}    & 24.24 & 0.06  & 10.03 & 10.01\\
                                              & \textbf{Ours}   & 0.91 & \textbf{24.80} & \textbf{0.06} & \textbf{5.44} & \textbf{4.03}\\ \midrule
\multirow{5}{*}{CATERv1}                      & MAGE$^{\dagger}$   & 0.97 & 35.03& 0.20  & 40.09 & 20.75\\
                                              & TiV-ODE         & 0.96 & -    & 0.12  & 11.98 & - \\
                                              & MAGE+$^{\dagger}$ & 0.96 & 36.16 & 0.09 & 6.41 & 21.15\\
                                              & Seer & 0.90 & 20.49 & 0.40 & 95.07 & 647.39 \\
                                              & \textbf{Ours} & \textbf{0.97} & \textbf{37.15} & \textbf{0.06} & \textbf{6.06} & \textbf{13.82}\\ \midrule
\multirow{5}{*}{CATERv2}                      & MAGE$^{\dagger}$   & 0.95 &32.54 & 0.22  & 28.16 & 51.81\\
                                              & TiV-ODE& 0.93 & -    & 0.18  & 38.12 & -\\
                                              & MAGE+$^{\dagger}$ & 0.95 & 32.58 & 0.11 & 23.56 & 34.93 \\
                                              & Seer & 0.86 & 18.51 & 0.42 & 79.31 & 589.74 \\
                                              & \textbf{Ours}   &\textbf{0.95} &\textbf{32.66} &\textbf{0.09} &\textbf{6.67} & \textbf{16.48} \\ \bottomrule
\end{tabular}
\caption{Quantitive comparison of our model and other models. $^{\dagger}$ means reproducing the results.}
\label{tab:Quantitive Results}
\end{table}
\subsection{Quantitive Results}
We compare TIV-Diffusion with several state-of-the-art models using SSIM, PSNR, LPIPS, FID, and FVD metrics. Throughout both the training and testing phases, objects in the video move at random speeds. Leveraging a given text caption and initial image, we generate nine subsequent frames, excluding the initial image from quantitative metric calculations. To maintain consistency, we reproduce MAGE's~\cite{hu2022MAGE} results on slightly modified MNIST datasets. Additionally, for the CATER datasets, we replicate the results of MAGE and MAGE+~\cite{hu2023MAGE+} using official weights to ensure uniform test conditions. We follow the fine-tuning procedures of Seer~\cite{gu2023seer} to train and test on the CATER datasets. Quantitative results for TiV-ODE and TVP are sourced from ~\cite{xu2023TiV-ODE} and ~\cite{song2022TVP}, respectively, and are summarized in Tab. \ref{tab:Quantitive Results}.

Our method demonstrates superiority in FID, FVD, and LPIPS metrics while remaining competitive in SSIM and PSNR compared to other models. The powerful generation capabilities of diffusion models have significantly heightened the perceptual quality of the resulting videos. Additionally, we enhance the alignment between text captions and images through object disentanglement, ensuring semantic consistency. The mitigation of issues related to object deformation or disappearance further improves visual quality of the generated video. These results collectively underscore the competitiveness of TIV-Diffusion for TI2V tasks.
\begin{table}[!h]
\centering
\small
\setlength{\tabcolsep}{1mm} 
\begin{tabular}{@{}cl|ccccc@{}}
\toprule
Datasets                                      & Method   & SSIM$\uparrow$ & PSNR$\uparrow$ & LPIPS$\downarrow$ & FID$\downarrow$ & FVD$\downarrow$\\ \midrule
\multirow{2}{*}{Single}                       & Ours w/o & 0.99 & 34.50& 0.01  & 1.66 & 0.52\\
                                              & Ours     &\textbf{0.99} &\textbf{36.68} &\textbf{0.01} &\textbf{0.49} & \textbf{0.11} \\ \midrule
\multirow{2}{*}{Double}                       & Ours w/o & 0.96 & 31.06& 0.03  & 1.55 & 0.51\\
                                              & Ours     &\textbf{0.96} &\textbf{32.99} &\textbf{0.03} &\textbf{1.04} & \textbf{0.47}\\ \midrule
\multirow{2}{*}{Modified}                     & Ours w/o & 0.90 & 24.34& 0.06  & 6.37 & 6.47\\
                                              & Ours     &\textbf{0.91} &\textbf{24.80} &\textbf{0.06} &\textbf{5.44} & \textbf{4.03}\\ \midrule
\multirow{2}{*}{CATER-v1}                     & Ours w/o & 0.97 & 36.59& 0.07  & 6.99 & 30.87\\
                                              & Ours     &\textbf{0.97} &\textbf{37.15} &\textbf{0.06} &\textbf{6.06} & \textbf{13.82}\\ \midrule
\multirow{2}{*}{CATER-v2}                     & Ours w/o & 0.95 & 32.06 &0.09 &10.66  & 32.72\\
                                              & Ours     &\textbf{0.95} &\textbf{32.66}&\textbf{0.09}&\textbf{6.67} & \textbf{16.48}\\ \bottomrule
\end{tabular}
\caption{Quantitive results of the ablation study. ``Ours w/o" signifies the absence of object-centric representations, while ``Ours" indicates their presence.}
\label{tab:Ablation Results}
\end{table}
\subsection{Ablation Study}
\paragraph{Object Disentanglement.}
We train the encoder $Enc$ for object disentanglement on MNIST datasets and CATER datasets respectively. During its initial image processing, $Enc$ extracts slots, with each slot corresponding to an object. To facilitate clarity in our description, we visualize these slots using $Dec$. 
The object disentanglement effects on various datasets can be found in the \textbf{Supplementary Material~\ref{app:object disentanglement}}.

To demonstrate the advantages of object disentanglement, we compare the results with and without using object-centric representations, as shown in Tab.~\ref{tab:Ablation Results}, where object-centric textual-visual alignment enhances generation results. Qualitative demonstrations and additional ablation experiments are provided in the \textbf{Supplementary Material~\ref{sec:Qualitative Results of Ablation on Object Disentanglement}} and \textbf{\ref{sec:additional ablation study}}, respectively. We explore the impact of denoising timesteps on video quality and assess the contributions of major adopted blocks through ablation studies.
\section{Conclusion}
\label{sec:conclusion}
In this paper, we propose a model for Text-Image-to-Video Generation, named TIV-Diffusion. Our model extends the diffusion model to enable autoregressive video frame generation, reducing the computational resources required for training. In addition to incorporating the fused textual and visual knowledge with the scale-offset modulation, we introduce object-centric representations to improve cross-modal alignment. After the slot attention encoder processes the input image, we obtain the corresponding slot for each object, which contains object attribute information. Subsequently, we identify the target objects to be manipulated based on the accompanying text. This process parallels the way humans comprehend natural phenomena. To mitigate object disappearance or deformation, and given the stochastic nature of objects stored in slots, our model employs Gumbel-Softmax to fuse object attributes. Experimental results confirm that our model attains state-of-the-art performance on existing datasets.
\section*{Acknowledgments}
This work was supported in part by NSFC under Grant 623B2098, 62021001, and 62371434. This work was mainly completed before March 2024.
\bibliography{aaai25}
\clearpage
\newpage
\appendix
\section{Supplementary Material}
We present the generated results on MNIST in Sec.~\ref{sec:Generated Results on MNIST}. The qualitative results of ablation on object disentanglement are shown in Sec.~\ref{sec:Qualitative Results of Ablation on Object Disentanglement}. Sec.~\ref{sec:additional qualitative results} provides additional qualitative results, including coordinate system, object disentanglement, controllability evaluation, more generated samples, additional comparison results, qualitative results of Seer, and additional generated samples when an initial occlusion exists. Sec.~\ref{sec:implementation details} introduces the implementation details of TIV-Diffusion. In Sec.~\ref{sec:additional ablation study}, we provide supplementary ablation experiments to validate the effectiveness of the current configuration of our model. In Sec.~\ref{sec:Further analysis on SPADE}, we explore the impact of scale and offset in SPADE on object movement. We present the generation results on the more realistic Weizmann, KTH, and Bridge datasets in Sec.~\ref{sec:weizmann results}, Sec.~\ref{sec:kth results}, and Sec.~\ref{sec:bridge results}. We discuss the inference efficiency of our model in Sec.~\ref{sec:Inference Efficiency} and its generalization to other video tasks in Sec.~\ref{sec:Generalization to Other Video Tasks}. Sec.\ref{sec:Broader Impact} and Sec.~\ref{sec:limitations} discuss the Broader Impact and Limitations of this paper, respectively.
\subsection{Generated Results on MNIST}
\label{sec:Generated Results on MNIST}
\begin{figure}[!h]
   \centering
   \includegraphics[width=1.0\linewidth]{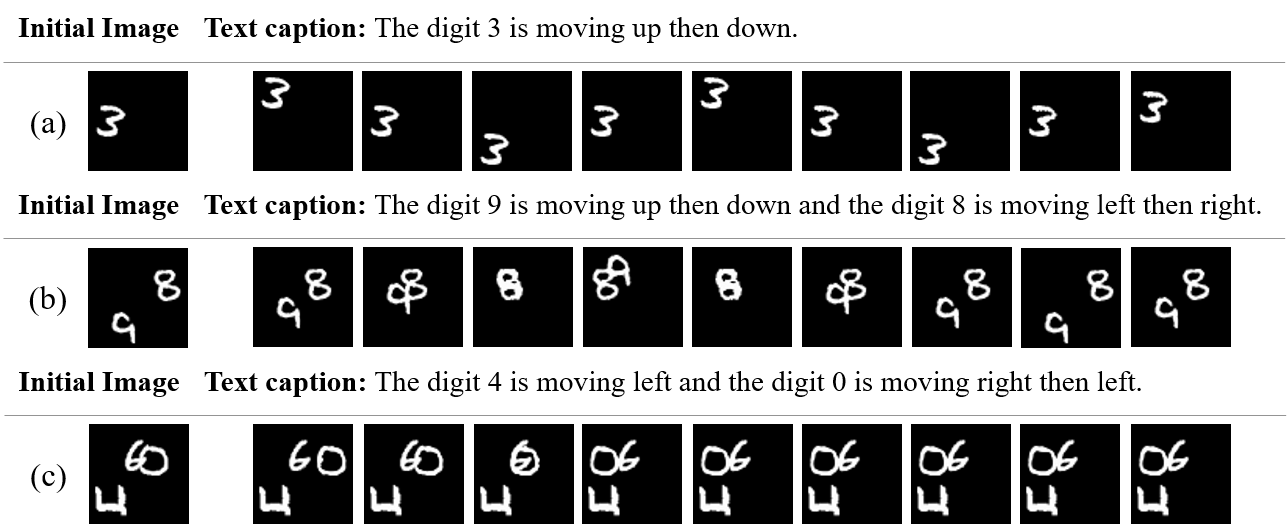}
   \caption{Samples generated from the MNIST datasets. (a) corresponds to Single, (b) corresponds to Double, and (c) corresponds to Modified.}
   \label{fig:mnist}
\end{figure}

Fig.~\ref{fig:mnist} depicts the generated results on the MNIST datasets, with the number of digits increasing from top to bottom. Our model consistently generates video frames that align with the given image content, ensuring clarity without blurring, and accurately matches the accompanying text caption. As the number of digits grows, overlapping may happen during movement, as seen in Fig.~\ref{fig:mnist}(b). Nonetheless, the results demonstrate that digit shapes are well-preserved. This is particularly evident in Fig.~\ref{fig:mnist}(c), where TIV-Diffusion accurately recognizes the digits despite the interference of the static digit. Concretely, the digit ``6'' remains stationary, ``0'' rebounds once, and ``4'' stays fixed at the left edge.
\begin{figure}[!h]
    \centering
    \includegraphics[width=1.0\linewidth]{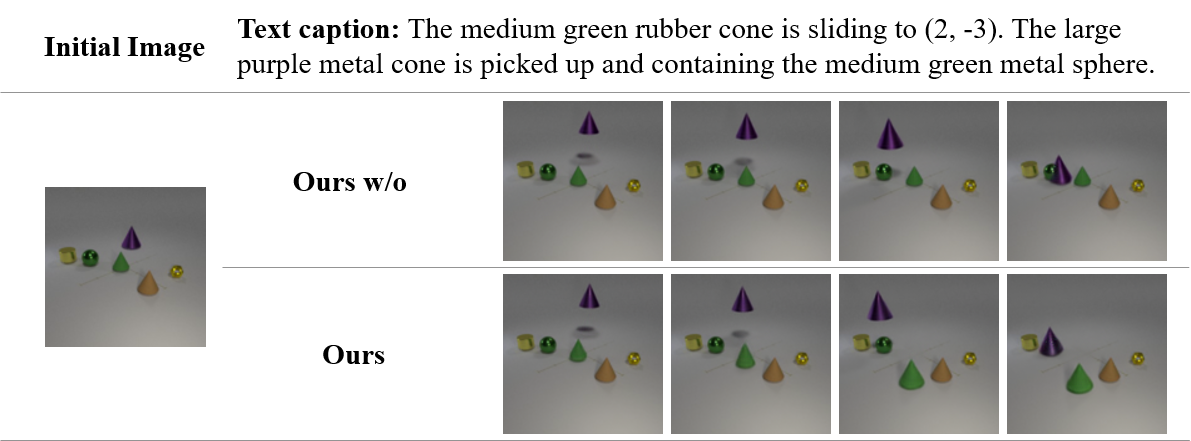}
    \caption{Comparison results of ``Ours" and ``Ours w/o". ``Ours w/o" denotes without using object-centric representations. The video generated by ``Ours" exhibits greater consistency with the accompanying text caption.}
    \label{fig:caterv2_ablation}
\end{figure}
\subsection{Qualitative Results of Ablation on Object Disentanglement}
\label{sec:Qualitative Results of Ablation on Object Disentanglement}
As shown in Fig.~\ref{fig:caterv2_ablation}, in the absence of object features, the model deviates when moving the purple metal cone, ultimately failing to fully contain the green sphere.
\subsection{Additional Qualitative Results}
\label{sec:additional qualitative results}
\subsubsection{Coordinate System}
\label{app:coordinate system}
\begin{figure}[!h]
    \centering
    \includegraphics[width=0.7\linewidth]{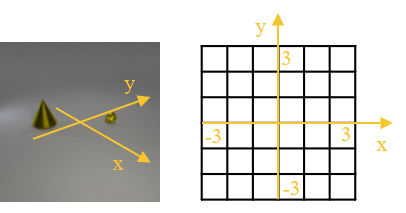}
    \caption{For CATER-GEN-v1, the coordinate system depicting the object locations is shown in the figure above. CATER-GEN-v2 adheres to the same coordinate system.}
    \label{fig:coordinate}
\end{figure}

CATER datasets share the same coordinate system, allowing users to instruct objects to move to specified coordinates through the text caption. A detailed illustration is provided in Fig.~\ref{fig:coordinate}.
\subsubsection{Object Disentanglement}
\label{app:object disentanglement}
\begin{figure}[!h]
    \centering
    \includegraphics[width=1.0\linewidth]{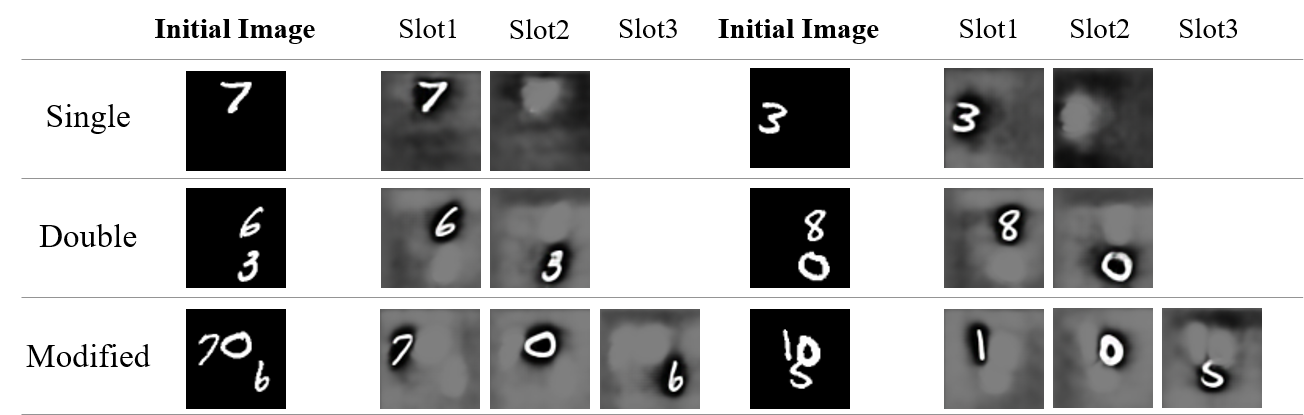}
    \caption{Object disentanglement results on MNIST datasets.}
    \label{fig:mnist_slots}
\end{figure}
\begin{figure}[!h]
    \centering
    \includegraphics[width=0.8\linewidth]{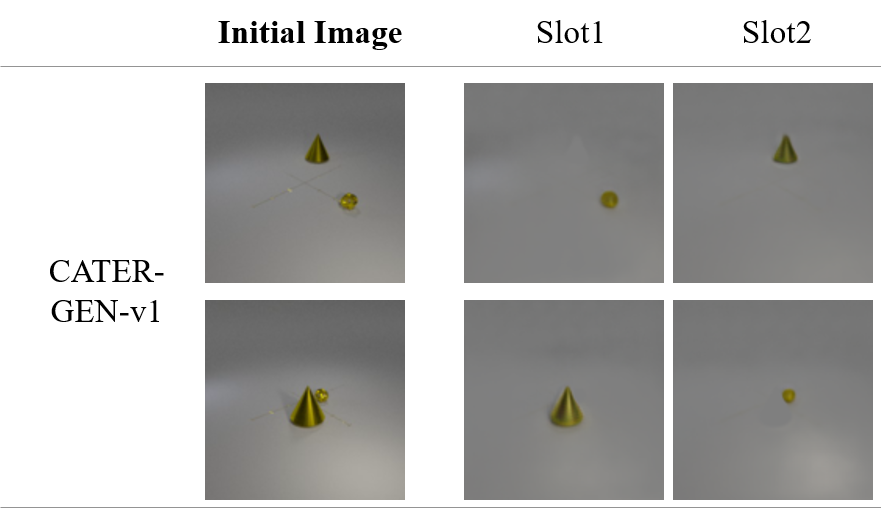}
    \caption{Object disentanglement results on CATER-GEN-v1.}
    \label{fig:caterv1_slots}
\end{figure}
\begin{figure}[!h]
    \centering
    \includegraphics[width=1.0\linewidth]{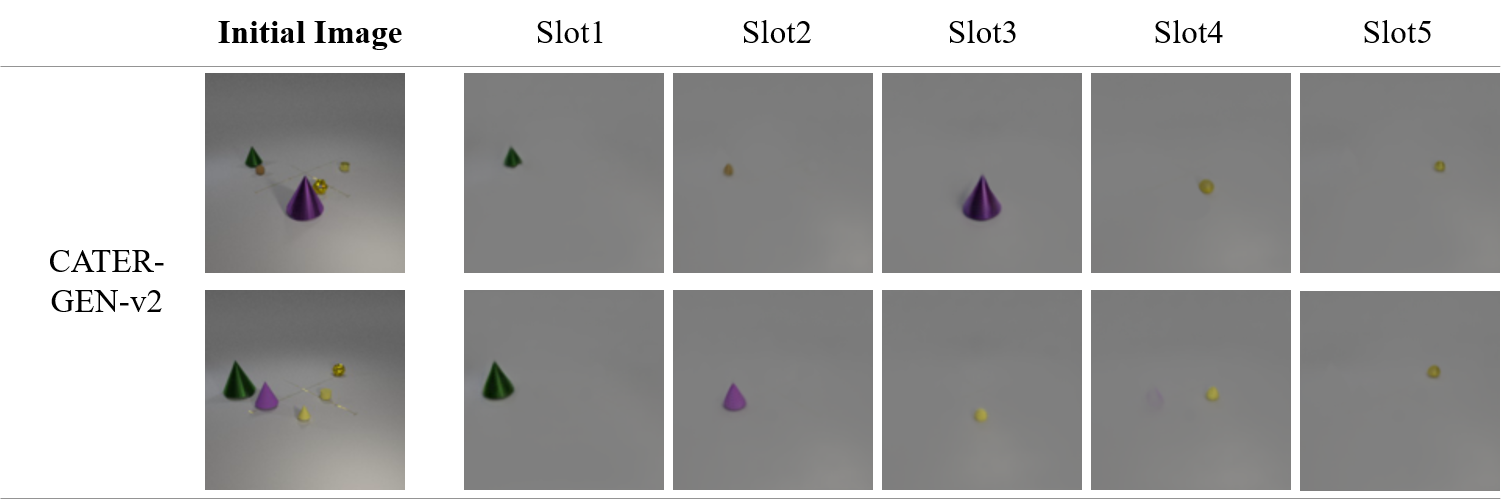}
    \caption{Object disentanglement results on CATER-GEN-v2. The Slot Attention encoder disentangles various objects within the initial image. Here, $Dec$ is employed for visualizing different slots.}
    \label{fig:cater_slots}
\end{figure}
Fig.~\ref{fig:mnist_slots} pertains to the MNIST datasets, where specific instances omit the slot corresponding to the background of objects for better visualization. It is evident that $Enc$ effectively discriminates between distinct objects. Even when a Single Moving MNIST image consists of only one digit, $Enc$ can distinguish the object from its background. Simultaneously, Fig.~\ref{fig:caterv1_slots} and Fig.~\ref{fig:cater_slots} illustrate the results of object disentanglement on CATER-GEN-v1 and CATER-GEN-v2.
\subsubsection{Controllability Evaluation}
\label{app:controllability evaluation}
\begin{figure*}[!t]
    \centering
    \includegraphics[width=1.0\linewidth]{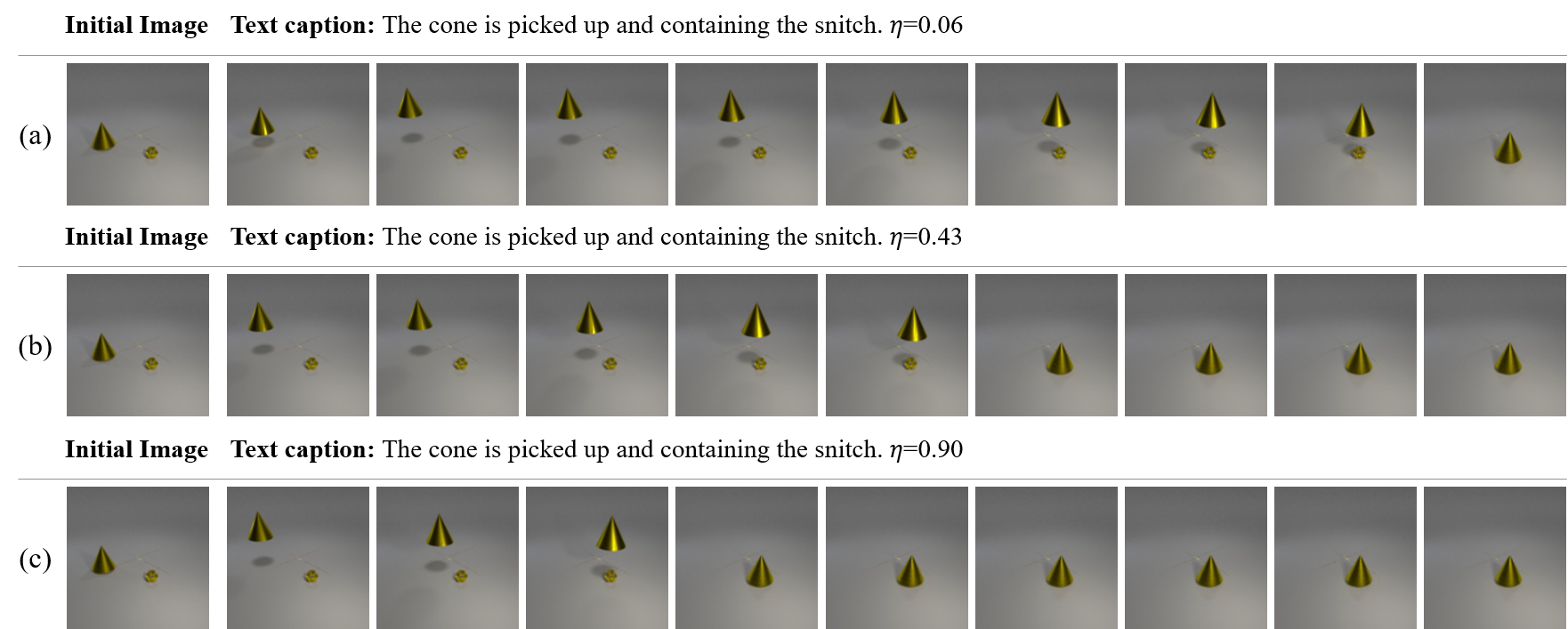}
    \caption{Samples generated from CATER-GEN-v1 for controllability evaluation. The initial image and text caption are the same, but the objects move at different speeds.}
    \label{fig:diverse speed}
\end{figure*}

\begin{figure*}[!t]
    \centering
    \includegraphics[width=1.0\linewidth]{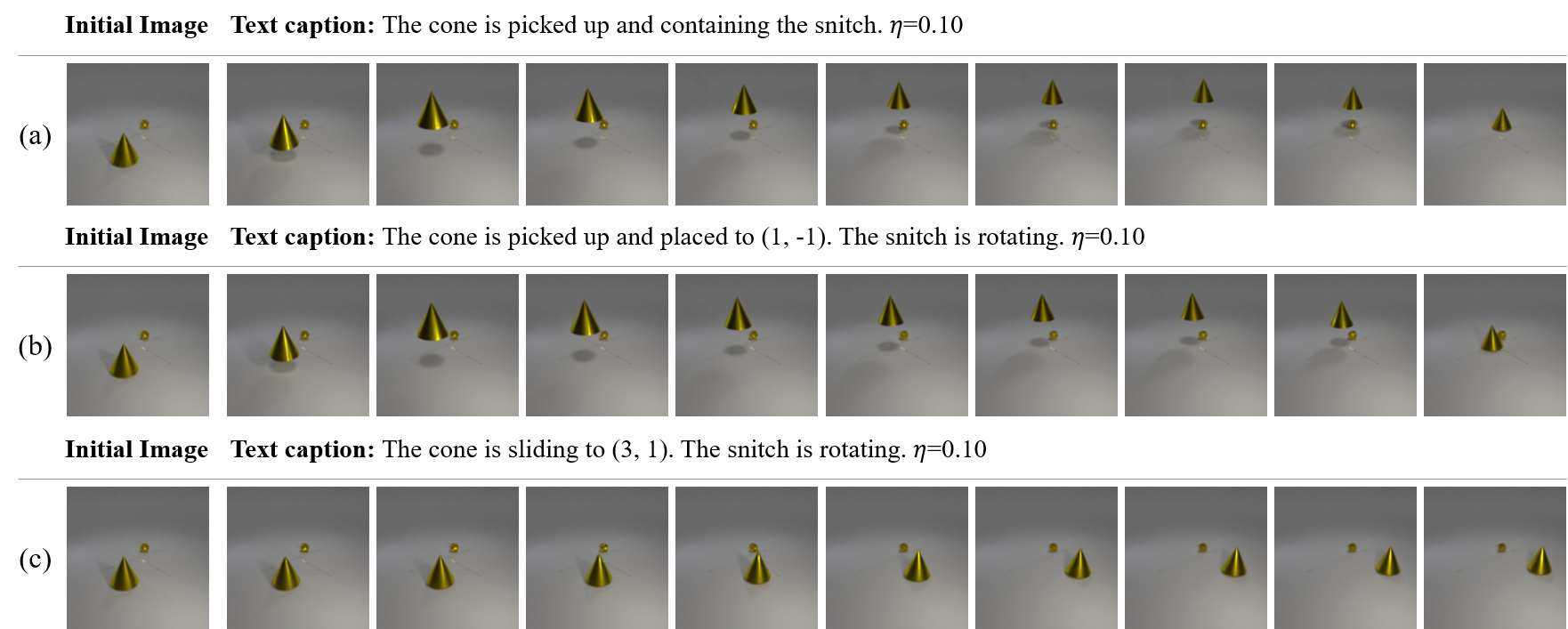}
    \caption{Samples generated from CATER-GEN-v1 for controllability evaluation. The initial image and speed are the same, but each line has a different text caption.}
    \label{fig:diverse text}
\end{figure*}

We conduct controllability evaluation on CATER-GEN-v1. Users can control the speed of object movement at will and specify any object to move to the desired location. For ease of presentation, we list two cases: fixed text caption but random speed and fixed speed but different text caption. The details are as shown in Fig.~\ref{fig:diverse speed} and Fig.~\ref{fig:diverse text}.

Fig.~\ref{fig:diverse speed} presents results for the same text caption but at different speeds on CATER-GEN-v1. As the value of $\eta$ increases, the cone's speed in containing the snitch accelerates, accompanied by natural changes in lighting and shadows on the object's surface. We showcase results using the same speed but different text captions to emphasize TIV-Diffusion's controllability further. Fig.~\ref{fig:diverse text} illustrates the model's proficiency not only in accurately locating the required objects but also in precisely moving them to specified positions. TIV-Diffusion adeptly captures subtle differences among various coordinates and affirms the effectiveness of its ``rotation" through changes in illumination on the snitch's surface.
\subsubsection{Additional Generated Samples}
\begin{figure*}[!t]
    \centering
    \includegraphics[width=1.0\linewidth]{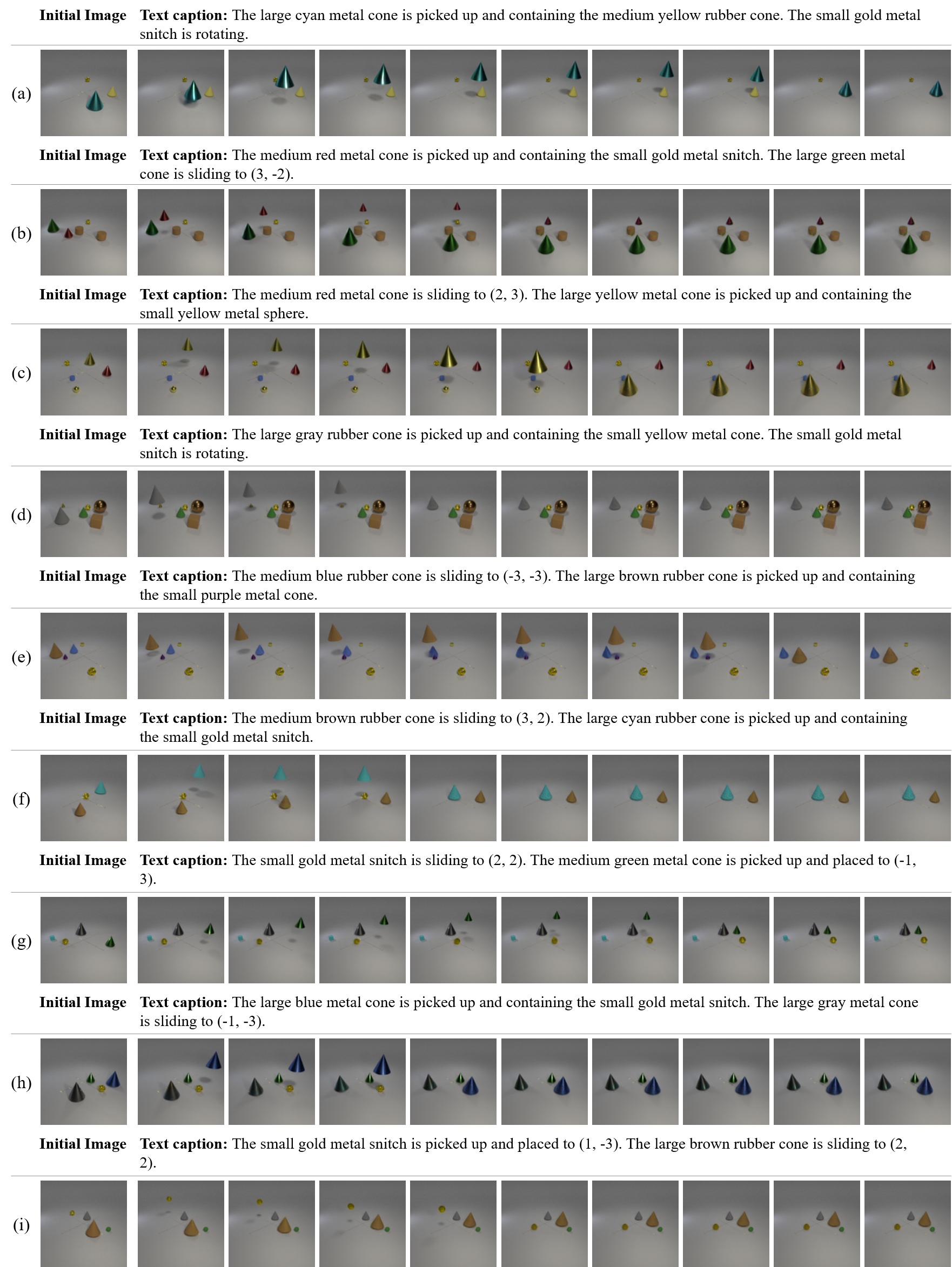}
    \caption{Additional samples generated from CATER-GEN-v2.}
    \label{fig:CATERv2 supp}
\end{figure*}

\begin{figure*}[!t]
    \centering
    \includegraphics[width=1.0\linewidth]{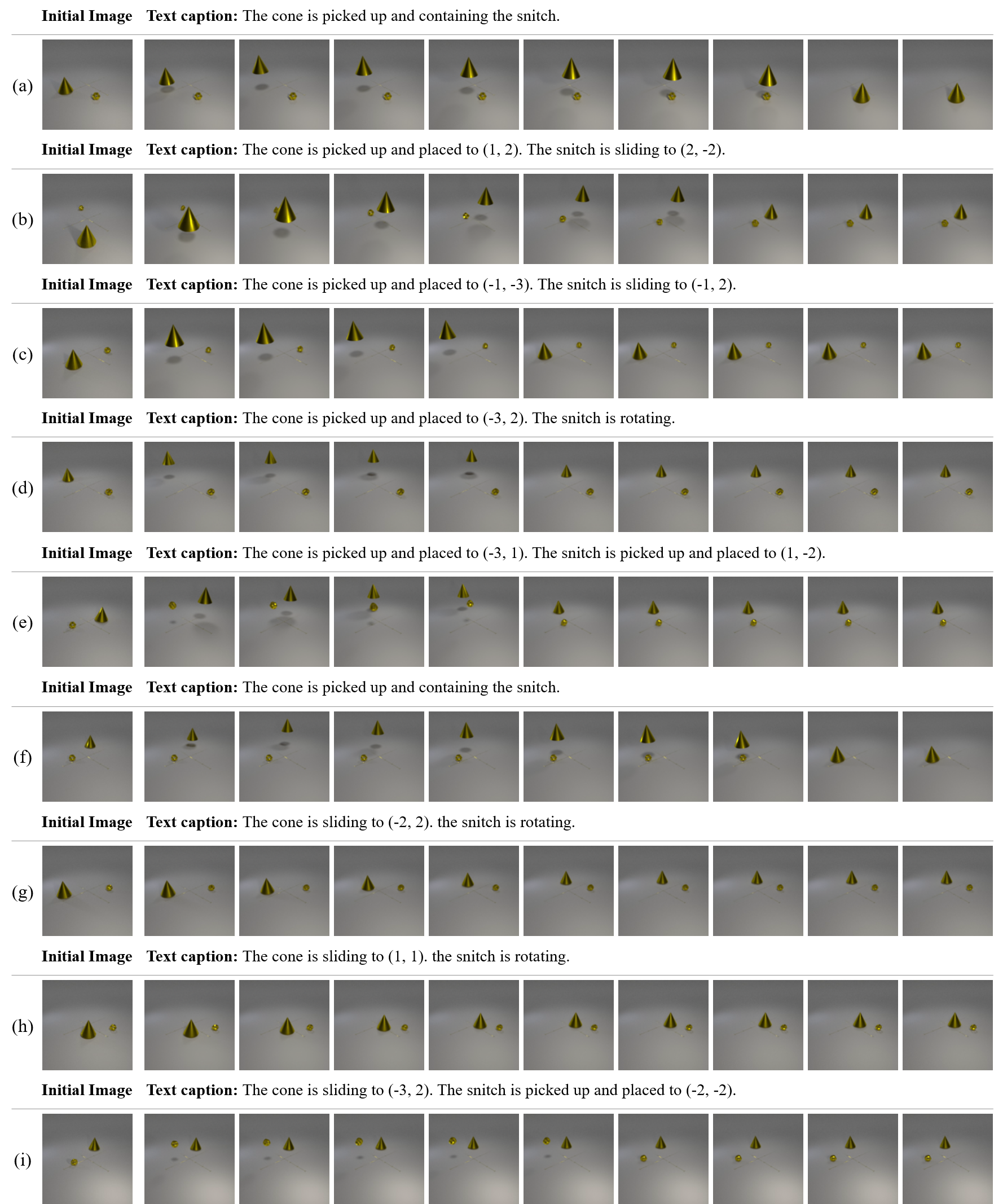}
    \caption{Additional samples generated from CATER-GEN-v1.}
    \label{fig:CATERv1 supp}
\end{figure*}

\begin{figure*}[!t]
    \centering
    \includegraphics[width=0.7\linewidth]{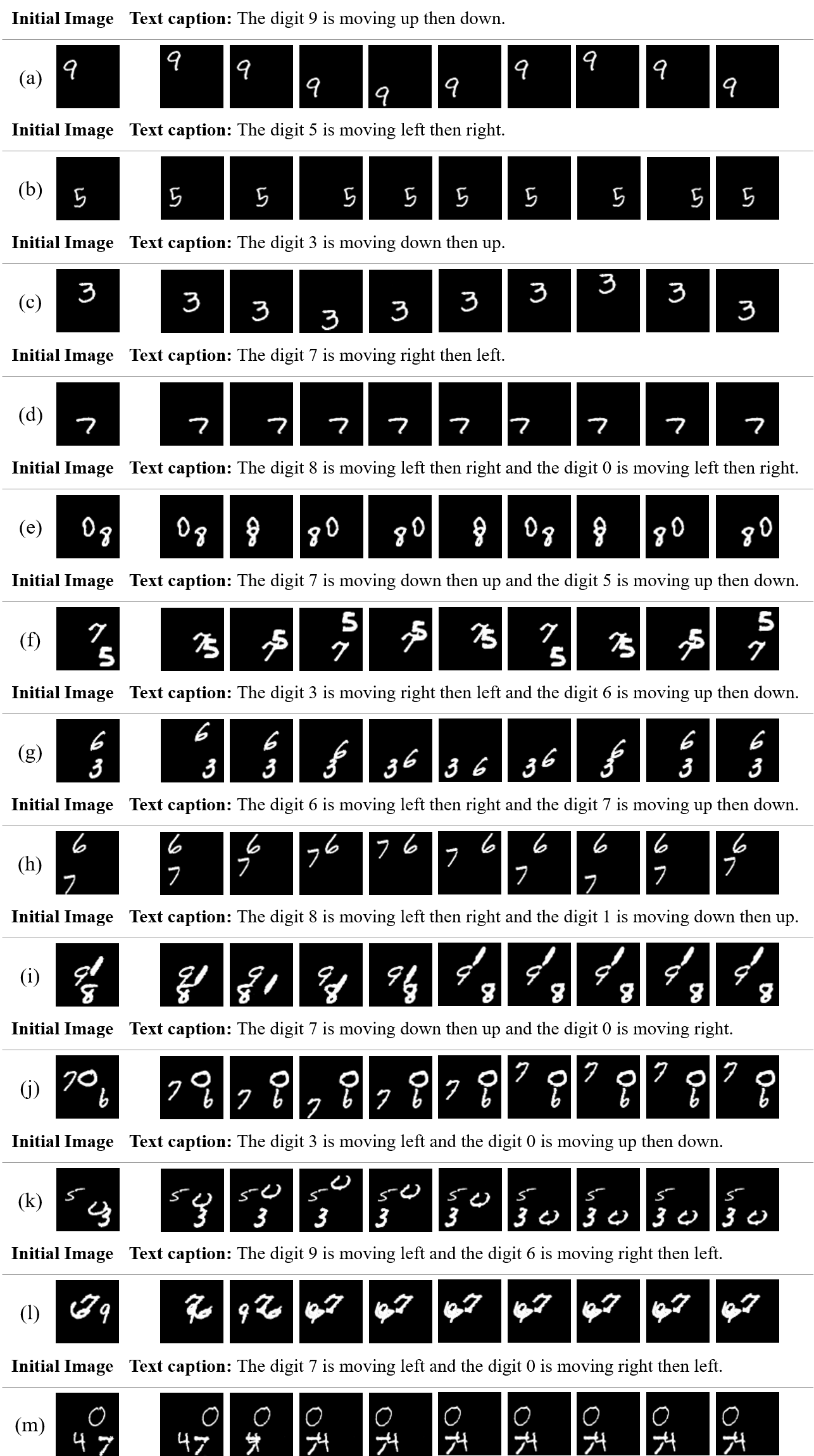}
    \caption{Additional samples generated from MNIST datasets.}
    \label{fig:MNIST supp}
\end{figure*}

We provide additional generated samples on CATER datasets of our TIV-Diffusion in Fig.~\ref{fig:CATERv2 supp} and Fig.~\ref{fig:CATERv1 supp}. Furthermore, more visualization of MNIST datasets is depicted in Fig.~\ref{fig:MNIST supp}.
\subsubsection{Additional Comparison Results}
\begin{figure*}[!t]
    \centering
    \includegraphics[width=1.0\linewidth]{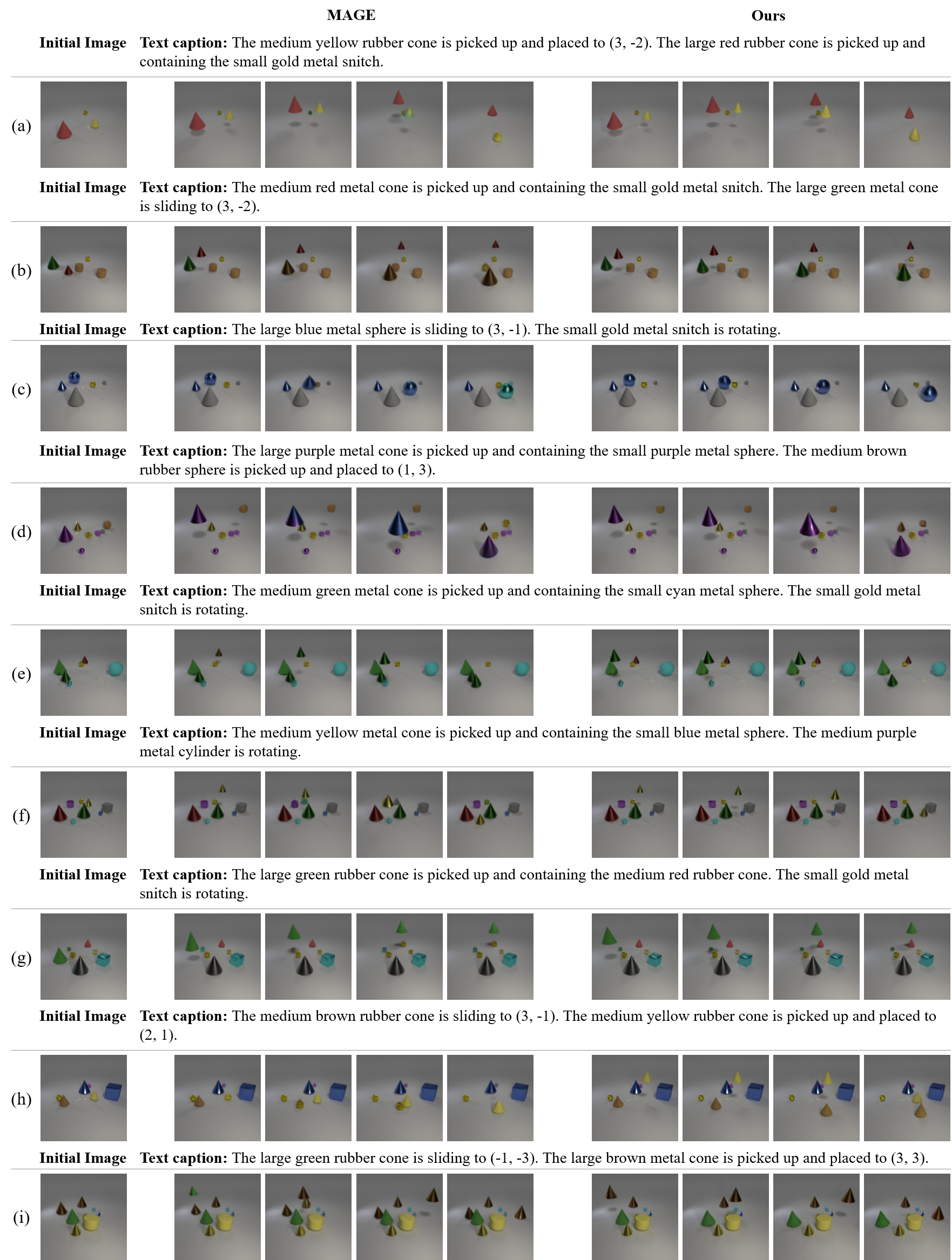}
    \caption{Additional comparison results on CATER-GEN-v2.}
    \label{fig:CATERv2 compa supp}
\end{figure*}

\begin{figure*}[!t]
    \centering
    \includegraphics[width=0.8\linewidth]{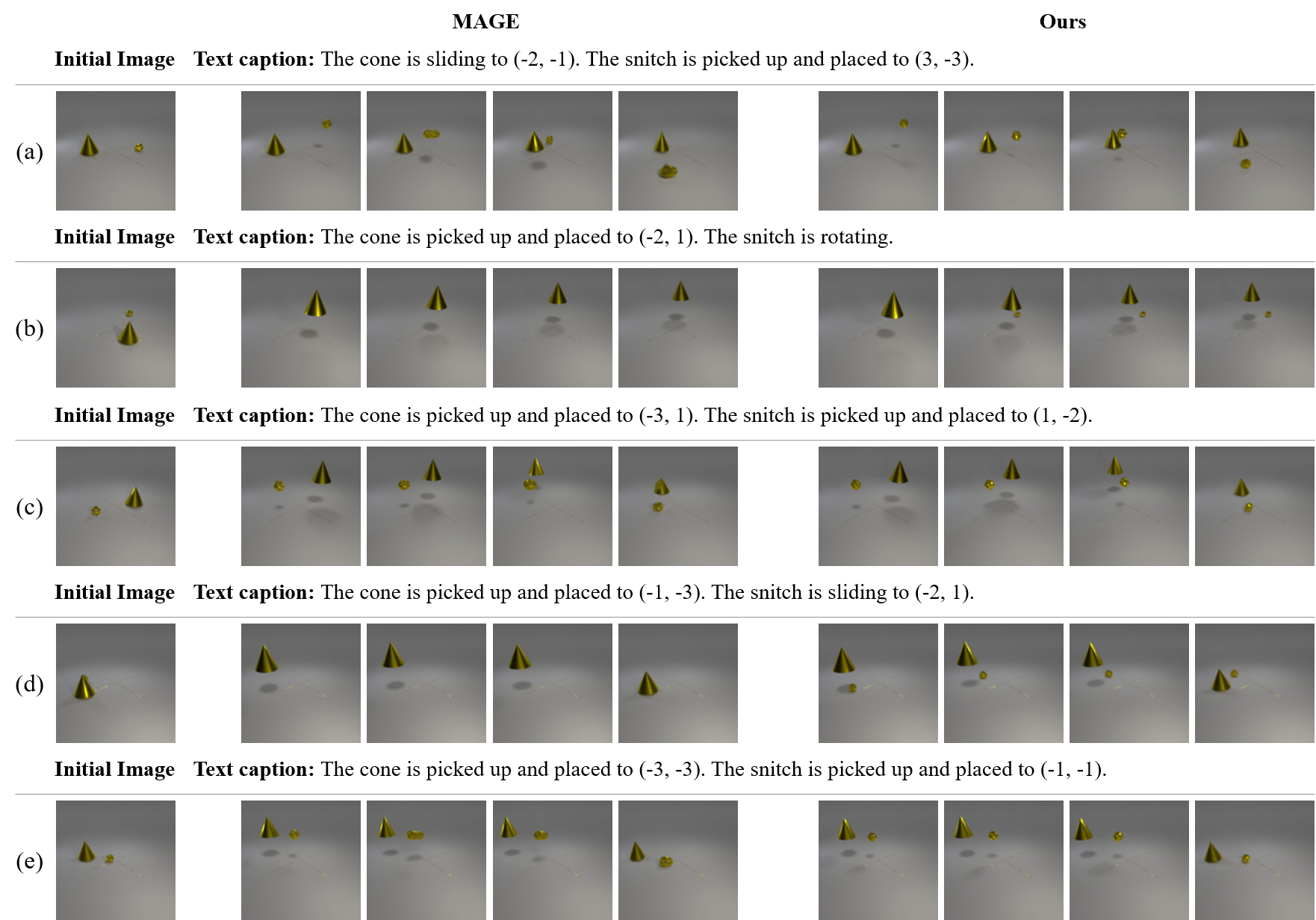}
    \caption{Additional comparison results on CATER-GEN-v1.}
    \label{fig:CATERv1 compa supp}
\end{figure*}

\begin{figure*}[!h]
    \centering
    \includegraphics[width=0.8\linewidth]{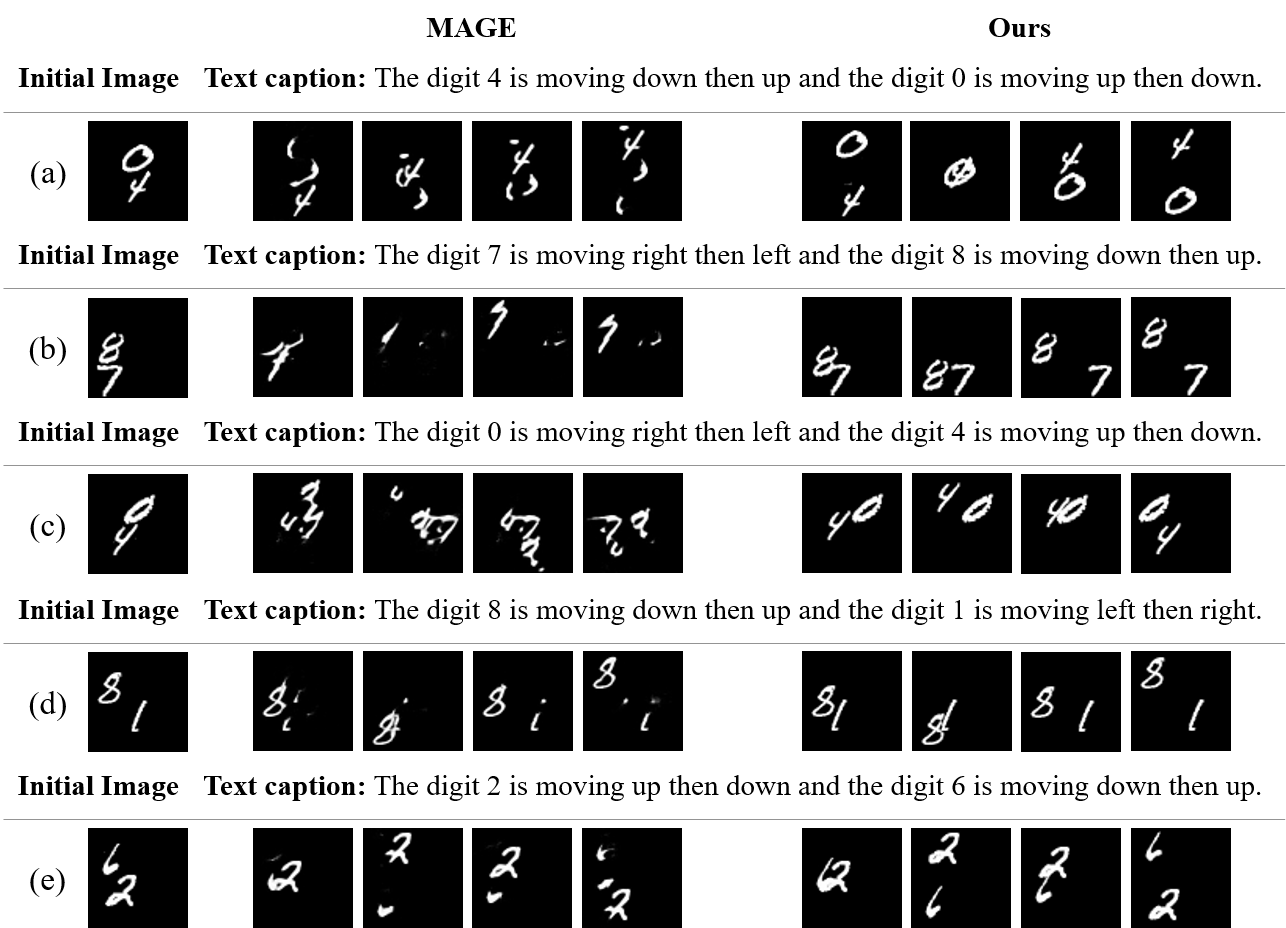}
    \caption{Additional comparison results on Double Moving MNIST.}
    \label{fig:Double compa supp}
\end{figure*}

TIV-Diffusion leverages object features to augment the semantic coherence between the generated video and textual caption while mitigating issues related to object deformation and disappearance. Further comparisons are presented in Fig.~\ref{fig:CATERv2 compa supp}, Fig.~\ref{fig:CATERv1 compa supp}, and Fig.~\ref{fig:Double compa supp}.
\subsubsection{Qualitative Results of Seer on CATER-GEN-v2}
\label{sec:Qualitative Results of Seer on CATERv2}
\begin{figure*}[h!]
    \centering
    \includegraphics[width=0.9\textwidth]{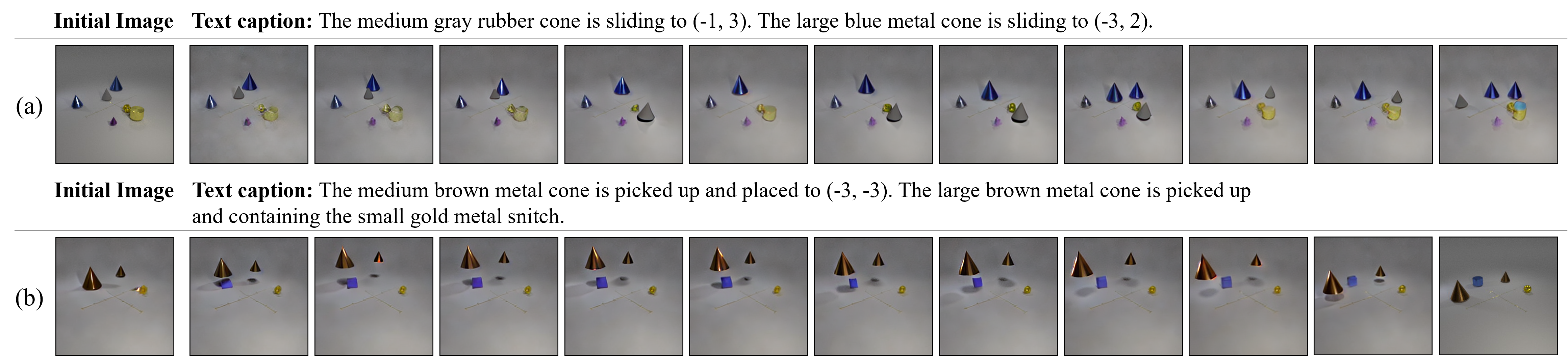}
    \caption{Qualitative results of Seer on CATER-GEN-v2.}
    \label{fig:Qualitative Results of Seer on CATERv2}
\end{figure*}

To illustrate the improvements in video quality of our method compared to generating all frames at once, we present the qualitative results of Seer on CATER-GEN-v2 in Fig.~\ref{fig:Qualitative Results of Seer on CATERv2}. As observed, Seer struggles to accurately follow the text descriptions and unexpectedly generates non-existent objects.
\subsubsection{Additional Generated Samples When An Initial Occlusion Exists}
\label{sec:initial occlusion}
\begin{figure*}
    \centering
    \includegraphics[width=0.8\linewidth]{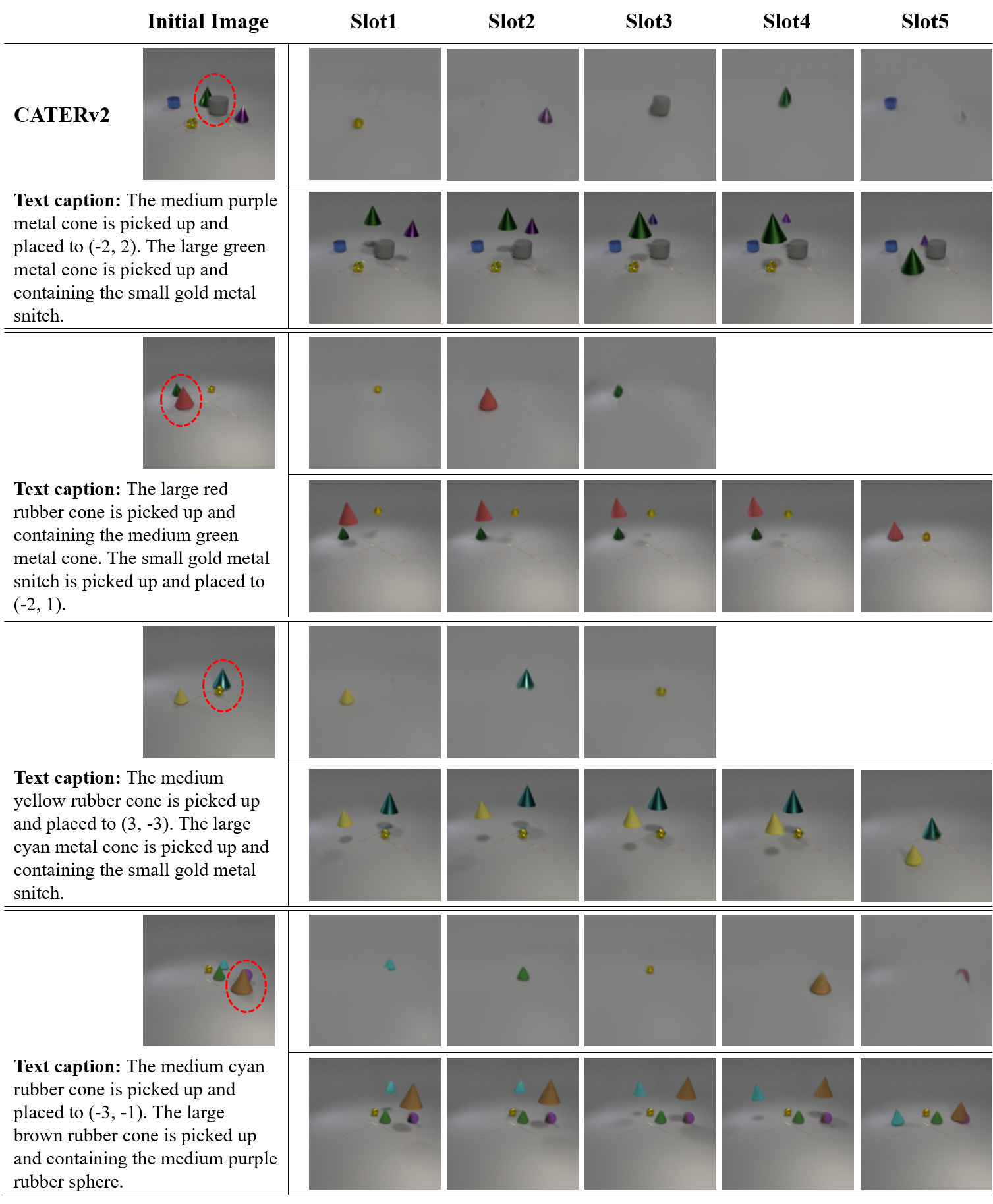}
    \caption{Additional generated samples when initial occlusion exists.}
    \label{fig:occlusion supp}
\end{figure*}

\begin{figure*}[!t]
    \centering
    \includegraphics[width=0.9\textwidth]{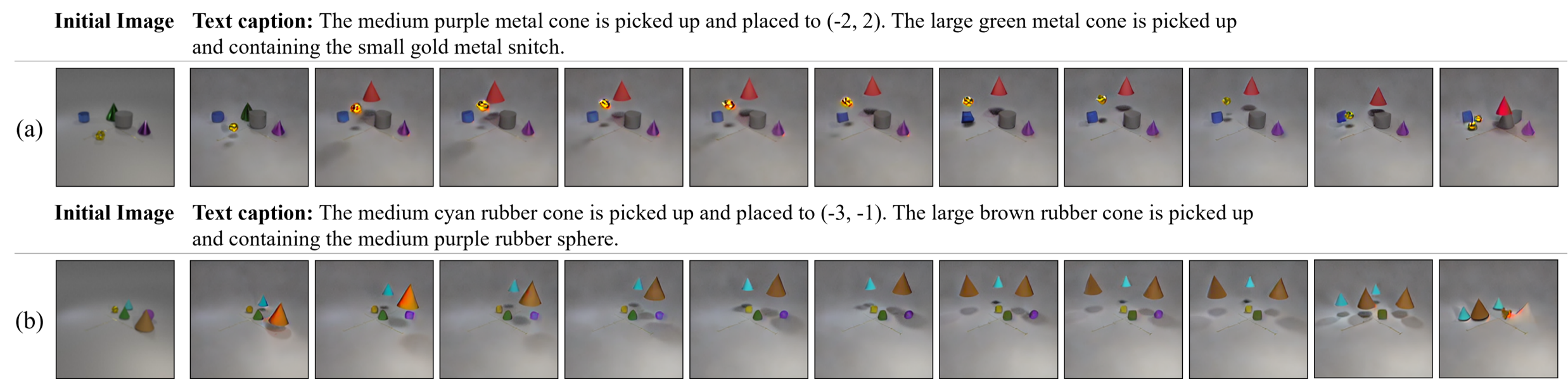}
    \caption{Seer's generation results with initial occlusion.}
    \label{fig:Seer's generation results with initial occlusion.}
\end{figure*}

We provide additional generated samples in Fig.~\ref{fig:occlusion supp} for scenarios with initial occlusion between objects in the input image on CATER-GEN-v2. Besides illustrating the movement process of objects guided by text captions, we also present object disentanglement results for the initial image. We compare the generated results of Seer under the same conditions in Fig.~\ref{fig:Seer's generation results with initial occlusion.} to further demonstrate the effectiveness of object disentanglement.
\subsection{Implementation Details}
\label{sec:implementation details}
All modules within TIV-Diffusion are trained from scratch, with the Slot Attention encoder trained before the autoregressive diffusion framework.
Before processing with Transformer Encoder layers, the image is divided into patches, and the size of each patch is $16\times 16$. The patch embedding dimensionality is $512$, consistent with the text embedding dimensionality. The speed variable $\eta$ is randomly chosen from the interval $(0, 1)$, with higher values indicating faster object motion. This randomness in speed is incorporated into the training process, where the video length $N$ is fixed at $10$. For MNIST datasets, the dimensionality of $D_{\mathrm{slots}}$ is set to $512$, and the \textit{Denoiser} employs a Channel Multiplier of $(1, 2, 4, 8)$. In the case of CATER datasets, $D_{\mathrm{slots}}$ is configured with a dimensionality of 128, and the \textit{Denoiser} utilizes a Channel Multiplier of $(1, 1, 2, 2, 4, 4)$. The initial learning rate is set to $5e-5$, suitable for training on $1$ to $3$ NVIDIA GeForce RTX 3090 GPUs.
\subsection{Additional Ablation Study}
\label{sec:additional ablation study}
\subsubsection{Denoising Timesteps} 
We evaluate the generated video quality across various denoising timesteps $T$ using CATER-GEN-v1, as shown in Tab.~\ref{tab:SamplingStep}. The perceptual quality consistently maintains a high standard, with only minor fluctuations in PSNR. This observation demonstrates the robustness of TIV-Diffusion under different denoising timesteps.
\begin{table}[!h]
\centering
\small
\setlength{\tabcolsep}{1mm}
\begin{tabular}{cc|c|c|c|c}
\toprule
\multicolumn{2}{c|}{Timesteps}                                      & $T$=250 & $T$=500  & $T$=1000 & $T$=1600\\ \midrule
\multicolumn{1}{c|}{\multirow{4}{*}{CATER-v1}}  & SSIM$\uparrow$    & 0.97 & 0.97  & 0.97   & \textbf{0.97}   \\ 
\multicolumn{1}{c|}{}                           & PSNR$\uparrow$    & 36.79 & 36.80 & 36.89  & \textbf{37.15}  \\
\multicolumn{1}{c|}{}                           & LPIPS$\downarrow$ & 0.06 & 0.06  & 0.06 & \textbf{0.06}   \\
\multicolumn{1}{c|}{}                           & FID$\downarrow$   & 6.41 & 6.41  & 6.42 & \textbf{6.06}    \\
\bottomrule
\end{tabular}
\caption{Different timesteps during sampling. TIV-Diffusion maintains high perceptual quality under different denoising timesteps.}
\label{tab:SamplingStep}
\end{table}

\subsubsection{Number of Slots}
Since the videos within CATER-GEN-v2 exhibit variability in the number of objects, we explore experimental performance across different slot counts, as illustrated in Tab.~\ref{tab:slots number}. Optimal results are achieved when $K=5$, and having too many or too few slots proves detrimental to the model's generated videos.
\begin{table}[!h]
\centering
\small
\begin{tabular}{cc|c|c|c|c}
\toprule
\multicolumn{2}{c|}{Slot Counts}                       & $K$=3      & $K$=5 & $K$=7 & $K$=11 \\ \midrule
\multicolumn{1}{c|}{\multirow{4}{*}{CATER-v2}}  & SSIM$\uparrow$    & 0.95 & \textbf{0.95}  & 0.95   & 0.95   \\ 
\multicolumn{1}{c|}{}                           & PSNR$\uparrow$    & 32.60& \textbf{32.66} & 32.39  & 32.16  \\
\multicolumn{1}{c|}{}                           & LPIPS$\downarrow$ & 0.09 & \textbf{0.09}  & 0.09   & 0.09\\
\multicolumn{1}{c|}{}                           & FID$\downarrow$   & 6.67 & \textbf{6.67}  & 6.72   & 6.81\\
\bottomrule
\end{tabular}
\caption{Number of slots. The optimal result is obtained when $K$=5.}
\label{tab:slots number}
\end{table}

\subsubsection{Different Conditional Scheme}
To further explore the impact of different conditional schemes on the generated videos, we replace the SPADE modulation method in TIV-Diffusion with the cross-attention mechanism and explore it on CATER datasets, as shown in Tab.~\ref{tab:Cross Attention}. The noisy frame serves as the Query, and the fused appearance and motion information is used as the Key and Value. In comparison, the SPADE method demonstrates more favorable results, prompting its selection as the default setting.

On the other hand, we substitute the Image Encoder in TIV-Diffusion with the Slot Attention encoder for image encoding, represented by ``Ours SA". 
The comparison results with the default setting ``Ours" are presented in Tab.~\ref{tab:Slot Attention Encoder}.
Relying solely on object disentanglement might result in information loss, thereby impacting the quality of the generated video. Similarly, ConvGRU captures video temporal relationships; replacing it with a convolutional layer will compromise object consistency. We present the ablation results of ConvGRU in Tab.~\ref{tab:ConvGRU}.
\begin{table}[!h]
\centering
\small
\begin{tabular}{@{}cl|cccc@{}}
\toprule
Datasets                                       & Method & SSIM$\uparrow$ & PSNR$\uparrow$ & LPIPS$\downarrow$ & FID$\downarrow$ \\ \midrule
\multirow{2}{*}{CATER-v1}                      & Ours CA  & 0.97 & 35.07 & 0.08  & 6.48\\
                                               & Ours  & \textbf{0.97} & \textbf{37.15} & \textbf{0.06}  & \textbf{6.06}\\
                                               \midrule
\multirow{2}{*}{CATER-v2}                      & Ours CA  & 0.93 & 29.23 & 0.14  & 9.47 \\
                                               & Ours  & \textbf{0.95} & \textbf{32.66} & \textbf{0.09}  & \textbf{6.67} \\
\bottomrule
\end{tabular}
\caption{Cross Attention. We replace the conditional scheme SPADE in TIV-Diffusion with a cross-attention mechanism, represented by ``Ours CA". ``Ours" is the default setting.}
\label{tab:Cross Attention}
\end{table}

\begin{table}[!h]
\centering
\small
\begin{tabular}{@{}cl|cccc@{}}
\toprule
Datasets                                       & Method & SSIM$\uparrow$ & PSNR$\uparrow$ & LPIPS$\downarrow$ & FID$\downarrow$ \\ \midrule
\multirow{2}{*}{CATER-v1}                      & Ours SA  & 0.97 & 35.29 & 0.07  & 7.08 \\
                                               & Ours  & \textbf{0.97} & \textbf{37.15} & \textbf{0.06}  & \textbf{6.06}\\
\bottomrule
\end{tabular}
\caption{Slot attention encoder. We substitute the Image Encoder in TIV-Diffusion with a convolutional layer, represented by ``Ours Conv". ``Ours" is the default setting.}
\label{tab:Slot Attention Encoder}
\end{table}
\begin{table}[!h]
\centering
\small
\setlength{\tabcolsep}{1mm}
\begin{tabular}{@{}cl|ccccc@{}}
\toprule
Datasets                                       & Method & SSIM$\uparrow$ & PSNR$\uparrow$ & LPIPS$\downarrow$ & FID$\downarrow$ & FVD$\downarrow$\\ \midrule
\multirow{2}{*}{CATER-v1}                      & Ours Conv  & 0.97 & 37.05 & 0.06  & 6.72 & 22.74\\
                                               & Ours  & \textbf{0.97} & \textbf{37.15} & \textbf{0.06}  & \textbf{6.06} & \textbf{13.82}\\
\bottomrule
\end{tabular}
\caption{Ablation results of ConvGRU. We substitute the ConvGRU in TIV-Diffusion with the Slot Attention encoder, represented by ``Ours SA". ``Ours" is the default setting.}
\label{tab:ConvGRU}
\end{table}
\subsection{Further Analysis on SPADE}
\label{sec:Further analysis on SPADE}
\begin{figure*}[!h]
    \centering
    \includegraphics[width=0.8\linewidth]{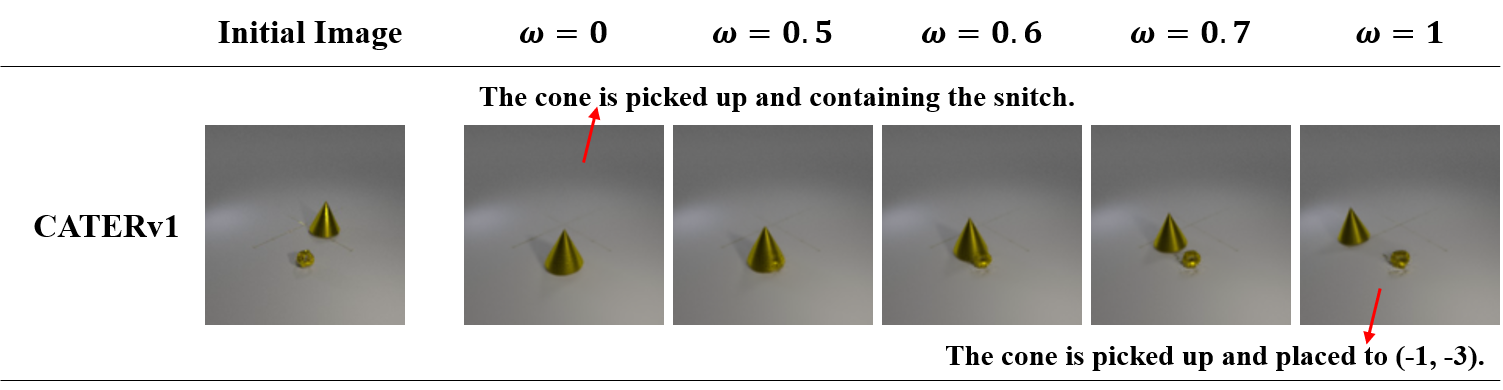}
    \caption{How scale in SPADE affects object movement.}
    \label{fig:rebuttal_Q3}
\end{figure*}
We first explore how scale in SPADE affects object movement using the CATERv1 dataset. Using two different text captions for the same initial image, we calculate the scale $\gamma _1$ and $\gamma _2$ with Eq. 2, then interpolate linearly to obtain $\gamma_{\mathrm{lerp}}$, that is, $\gamma _{\mathrm{lerp}}=\gamma _1+\omega \left( \gamma _2-\gamma _1 \right)$. The text caption for $\gamma _1$ states, ``The cone is picked up and containing the snitch,'' while the caption for $\gamma _2$ describes, ``The cone is picked up and placed to (-1, -3).'' We observe that by adjusting $\omega \in \left[ 0,1 \right]$, the object is moved to a position halfway between the locations specified by the two text captions, as shown in Fig. \ref{fig:rebuttal_Q3}. We similarly explore the offset and observe that it does not influence object movement, but marginally alters the brightness of the objects' surfaces.
\subsection{Results on the Weizmann Dataset}
\label{sec:weizmann results}
\begin{figure*}[h!]
    \centering
    \includegraphics[width=\textwidth]{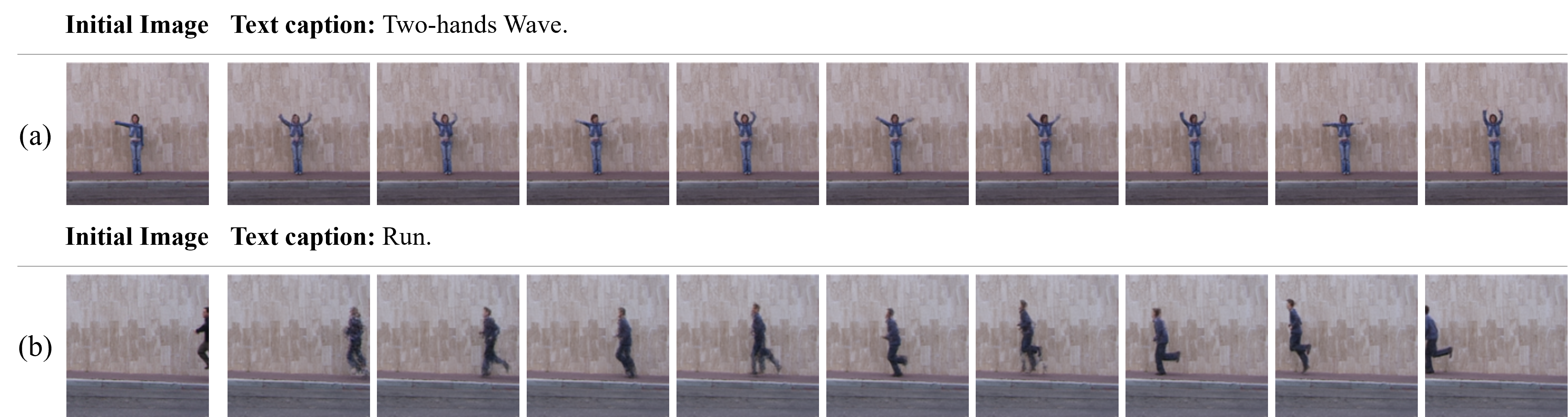}
    \caption{Qualitative Results of TIV-Diffusion on Weizmann Dataset.}
    \label{fig:Weizmann}
\end{figure*}
\paragraph{Weizmann dataset.} The Weizmann dataset~\cite{gorelick2007actions} contains 10 actions (walk, run, jump, gallop sideways, bend, one-hand wave, two-hands wave, jump in place, jumping jack, skip), with 9 different samples for each action. We split the dataset into an 8:2 training and testing ratio.
\paragraph{Generated Results.} We present the qualitative testing results in Fig.~\ref{fig:Weizmann}.
\subsection{Results on the KTH Dataset}
\label{sec:kth results}
\begin{figure*}[h!]
    \centering
    \includegraphics[width=\textwidth]{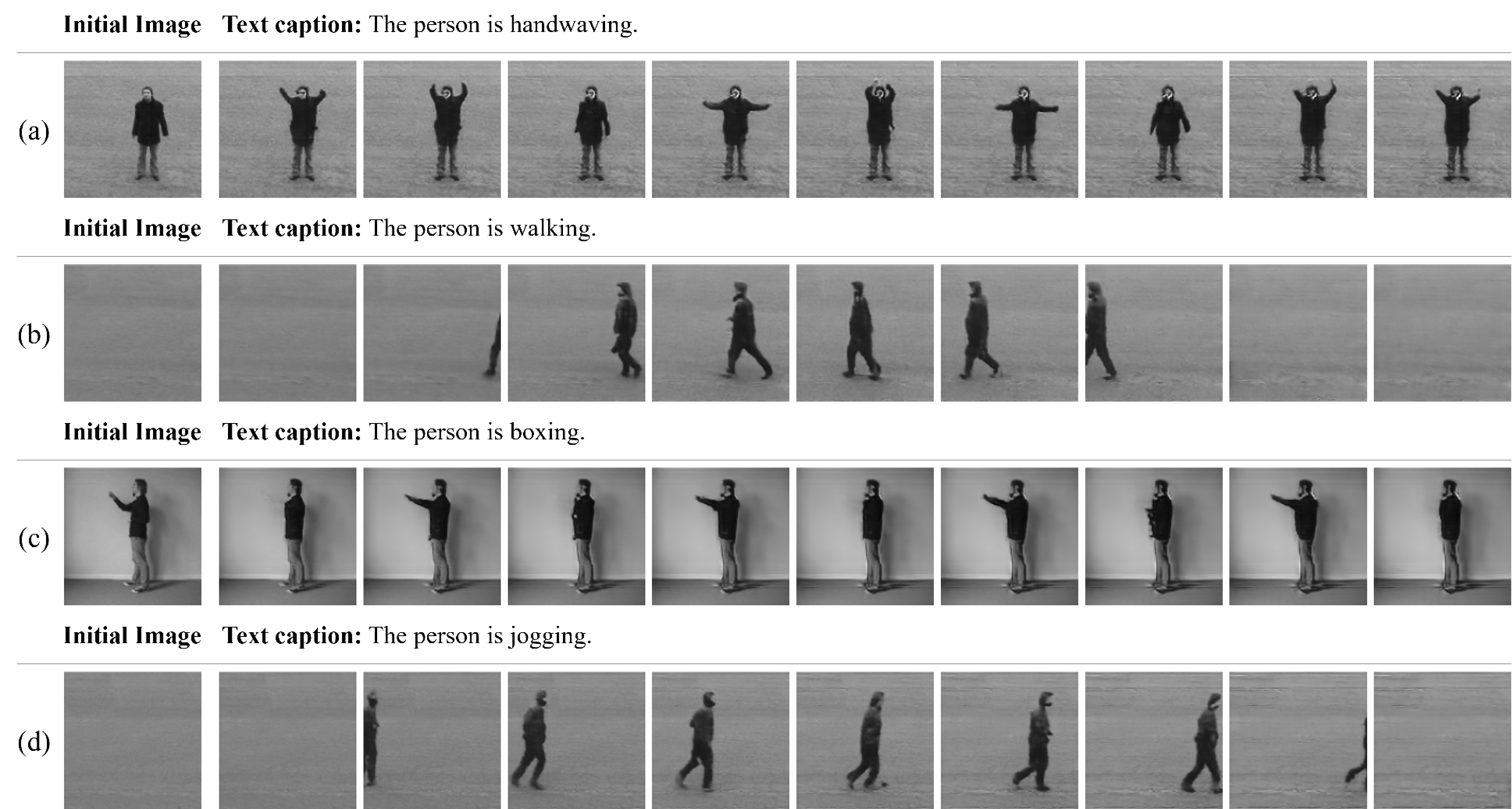}
    \caption{Qualitative Results of TIV-Diffusion on KTH Dataset.}
    \label{fig:kth}
\end{figure*}
\paragraph{KTH dataset.} The KTH dataset~\cite{schuldt2004kth} contains 2391 video clips of six human actions (walking, jogging, running, boxing, hand waving, and hand clapping) performed by 25 people in four different scenarios. We prepend ``The person is'' to each video's corresponding action label to form a complete text caption. We split the dataset into a training set and a testing set with an 8:2 ratio.
\paragraph{Generated Results.} Some generated samples can be seen in Fig.~\ref{fig:kth}.
\subsection{Results on the Bridge Dataset}
\label{sec:bridge results}
\begin{figure*}[!h]
    \centering
    \includegraphics[width=1.0\linewidth]
    {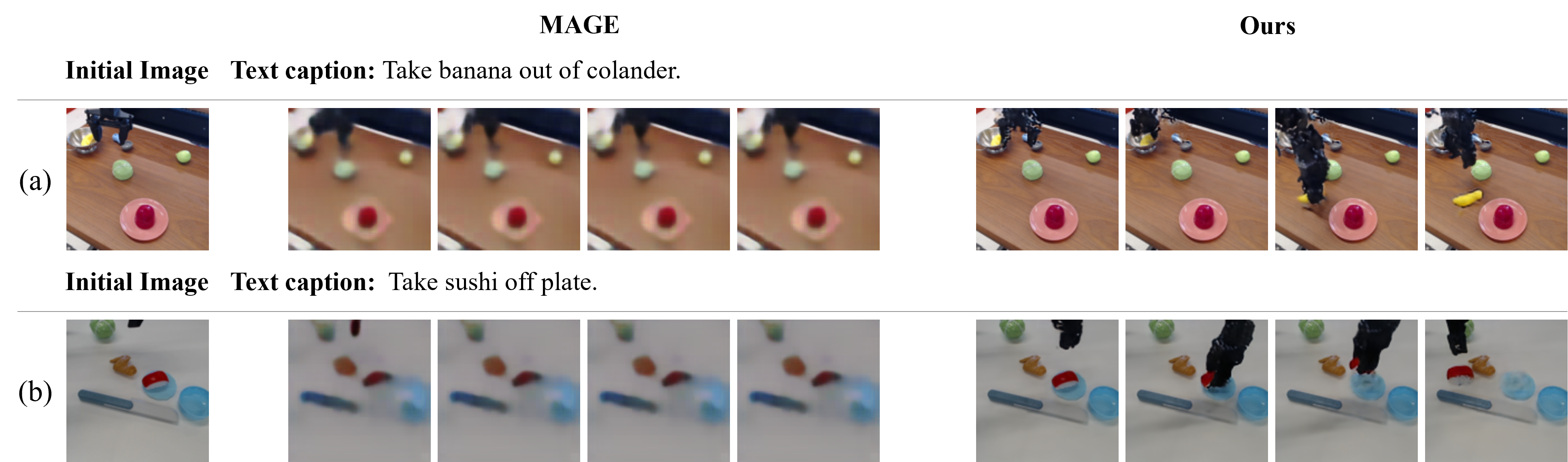}
    \caption{Comparison Results on Bridge.}
    \label{fig:bridge comparison}
\end{figure*}
\begin{figure*}[!h]
    \centering
    \includegraphics[width=1.0\linewidth]
    {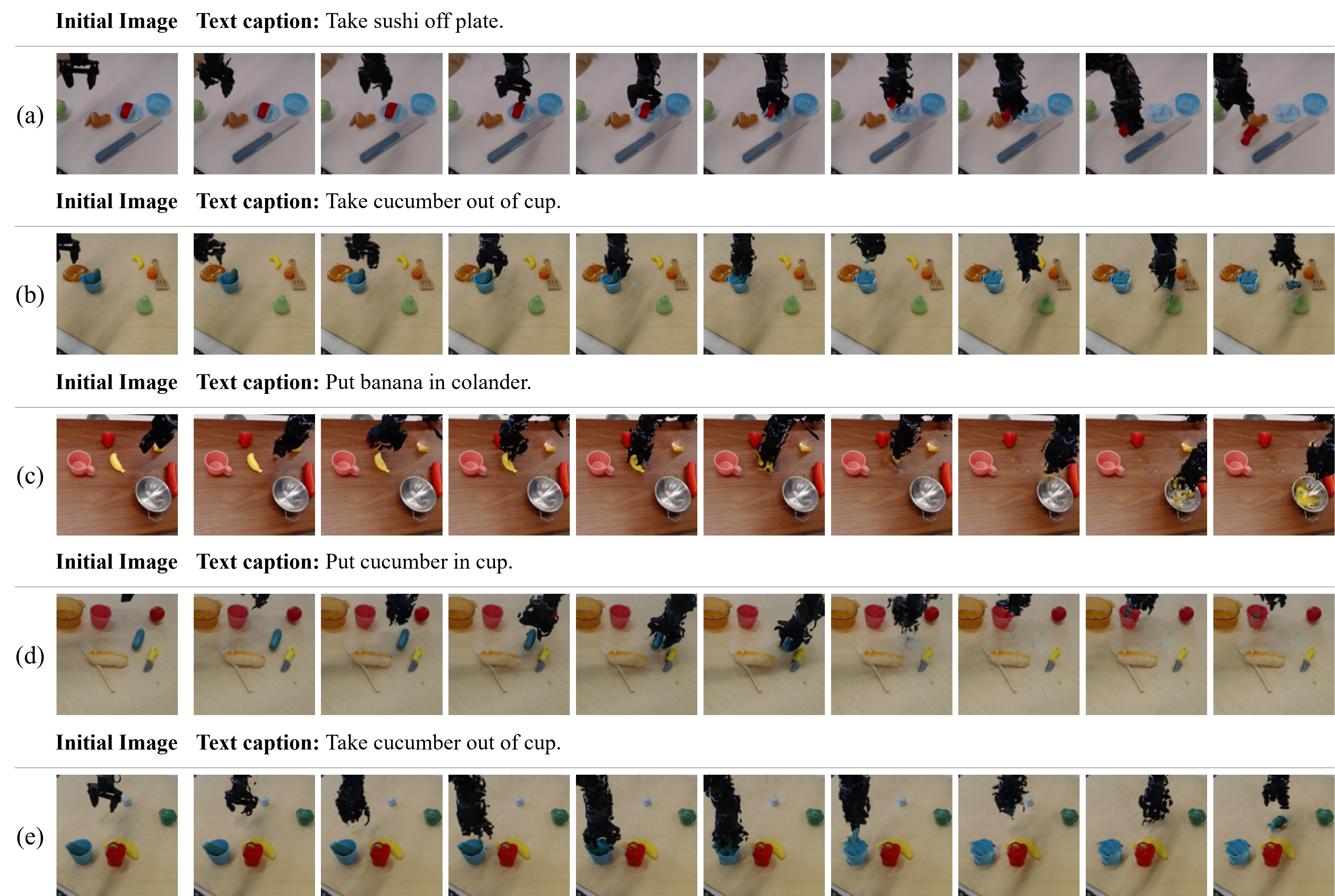}
    \caption{Samples generated from the Bridge dataset.}
    \label{fig:generated bridge}
\end{figure*}
\begin{table}[!h]
\centering
\small
\setlength{\tabcolsep}{1mm}
\begin{tabular}{@{}cl|ccccc@{}}
\toprule
Datasets                                       & Method & SSIM$\uparrow$ & PSNR$\uparrow$ & LPIPS$\downarrow$ & FID$\downarrow$ & FVD$\downarrow$\\ \midrule
\multirow{3}{*}{Bridge}                        & MAGE   & 0.65 & 16.77 & 0.51  & 179.09 & 1285.66\\
                                               & Seer   & 0.55 & 12.73 & 0.45  & 93.56  & 928.18\\
                                               & Ours   & \textbf{0.76} & \textbf{17.85} & \textbf{0.24}  & \textbf{59.18} & \textbf{326.36}\\
\bottomrule
\end{tabular}
\caption{Quantitative comparison of our model and other models on the Bridge dataset.}
\label{tab:Bridge dataset}
\end{table}
\paragraph{Bridge dataset.} The Bridge dataset~\cite{ebert2021bridge} is designed to enhance the generalization capabilities of robotic skills across different domains. The tasks in this dataset are primarily centered around kitchen-related activities, utilizing a low-cost WidowX250s robotic arm to perform various operations. These demonstrations were collected using an Oculus Quest2 VR device, with multiple perspectives captured by flexibly mounted cameras.
\paragraph{Comparison Results.} We compare the generation results of MAGE and TIV-Diffusion on the more realistic Bridge dataset. We select videos from three scenes: ``tabletop\_dark\_wood,'' ``tabletop\_light\_wood,'' and ``tabletop\_white.'' These data are split into training and testing sets with an 8:2 ratio. As shown in Fig.~\ref{fig:bridge comparison}, Our method not only maintains consistency with the initial image but also ensures that the robotic arm's movements follow the input commands. In contrast, MAGE generates blurred video frames and the robotic arm remains mostly static, failing to align with the text descriptions. Additionally, we quantitatively evaluate the generated results, as shown in Tab.~\ref{tab:Bridge dataset}. TIV-Diffusion outperforms MAGE across all metrics, demonstrating the effectiveness of our method on the Bridge dataset.
\paragraph{Generated Results.} In Fig.~\ref{fig:generated bridge}, we present some samples generated by TIV-Diffusion on the Bridge dataset. As illustrated, the initial image includes a robotic arm in random positions, varying table colors, and different objects on the table. Despite these variations, our method consistently performs the actions as instructed by the text.
\subsection{Inference Efficiency}
\label{sec:Inference Efficiency}
We assess the time required for single-frame generation at different inference timesteps and model complexity on CATERv1 on 1 NVIDIA GeForce RTX 3090 GPU, as shown in Tab.~\ref{tab:InferenceTime}. To minimize GPU requirements for video generation, we employ autoregressive generation. Although this may lead to increased inference time, it achieves better results compared to generating all video frames at once (\textit{e.g.}, Seer), without the need for expensive computational costs.
\begin{table}[!h]
\centering
\small
\setlength{\tabcolsep}{0.5mm}
\begin{tabular}{c|cccc|c|c}
\toprule
\multicolumn{1}{c|}{Timesteps}     & $T$=250 & $T$=500  & $T$=1000 & $T$=1600 & GFlops & Params\\ \midrule
\multicolumn{1}{c|}{CATERv1}      & 7$s$    & 15$s$    & 31$s$    & 53$s$    & 152.92 & 80.84$M$\\ 
\bottomrule
\end{tabular}
\caption{Inference time and computational complexity.}
\label{tab:InferenceTime}
\end{table}
\subsection{Generalization to Other Video Tasks}
\label{sec:Generalization to Other Video Tasks}
Our model can be extended to other tasks. The results currently presented can be viewed as video prediction tasks guided by text. If the goal is solely to perform video prediction, the branch that extracts text information and fuses it with the given image features can be removed, leaving the ConvGRU to capture the temporal relationships of the input video frames. Similarly, the model can be used for video interpolation tasks. In this case, we conduct corresponding experiments by conditioning the model on the first and last frames of the video, predicting the interpolation frames without text prompts. The processing pipeline for the first frame remains unchanged, while the last frame, after being encoded by the Image Encoder, replaces the text embedding in the original model for subsequent feature fusion. We train and test this setup on the CATERv1 dataset, and the quantitative results are presented in the Tab.~\ref{tab:frame interpolation}.
\begin{table}[!h]
\centering
\small
\setlength{\tabcolsep}{1mm}
\begin{tabular}{@{}cc|ccccc@{}}
\toprule
Datasets                                       & Method & SSIM$\uparrow$ & PSNR$\uparrow$ & LPIPS$\downarrow$ & FID$\downarrow$ & FVD$\downarrow$ \\ \midrule
\multirow{1}{*}{CATERv1}                      & Ours Inter  & 0.96 & 33.19 & 0.10  & 5.44  & 53.01\\
\bottomrule
\end{tabular}
\caption{Quantitative Results of our model for Frame Interpolation.}
\label{tab:frame interpolation}
\end{table}
\subsection{Broader Impact}
\label{sec:Broader Impact}
Text-driven Image to Video Generation (TI2V), as implemented in our TIV-Diffusion framework, leverages open-source datasets and avoids generating harmful content. By introducing an object-centric textual-visual alignment module, we improve the generation quality. TI2V's primary applications span entertainment, education, and digital content creation. Importantly, the deployment of TIV-Diffusion adheres to strict privacy and safety standards, underlining our commitment to responsible and ethical technological advancement.
\subsection{Limitations}
\label{sec:limitations}
While our study demonstrates significant advancements across various datasets, several limitations should be acknowledged. Our method has more parameters compared to MAGE, which leads to improved generation quality but also increases the model size. Additionally, the autoregressive strategy of TIV-Diffusion results in a longer inference time. The quality and diversity of the training data affect the generalization capability of the generated videos to unseen scenarios. Future work could explore the effectiveness of our approach in more diverse and complex environments.
\end{document}